\documentclass[10pt,onecolumn]{article}

\usepackage{chemfig}
\usepackage{float}
\usepackage{xr} 
\usepackage{hyperref}
\usepackage{gensymb}
\usepackage{lscape}
\usepackage{array,booktabs,longtable}
\newcolumntype{C}[1]{>{\centering\arraybackslash}m{#1}}
\usepackage{chemnum}
\usepackage{tabularx}
\usepackage{graphicx}
\usepackage{authblk}
\newcommand{\coo}{CO\textsubscript{2}}

\newcommand{\ndatafull}{130}
\newcommand{\ndata}{128}
\newcommand{\ntrain}{99}
\newcommand{\nval}{11}
\newcommand{\ntest}{20}

\newcommand{\codedataurl}{\url{https://github.com/IBM/Carbon-capture-fingerprint-generation}}
\newcommand{\ac}{absorption capacity}
\newcommand{\oir}{observed initial rate}
\newcommand{\si}{Supporting Information}

\newcommand{\gp}{Gaussian Process}
\newcommand{\ab}{Adaboost}
\newcommand{\svm}{Support Vector Machine}

\title{Machine Guided Discovery of Novel Carbon Capture Solvents}

\author[1,2]{James L. McDonagh}
\author[4]{Benjamin H. Wunsch}
\author[3]{Stamatia Zavitsanou}
\author[1]{Alexander Harrison}
\author[4]{Bruce Elmegreen}
\author[4]{Stacey Gifford}
\author[4]{Theodore van Kessel}
\author[1]{Flaviu Cipcigan}

\affil[1]{IBM Research Europe - UK, Hartree Centre, SciTech Daresbury, Warrington, Cheshire WA4 4AD, UK}
\affil[2]{Current address: Ladder Therapeutics doing business as Serna Bio, Lab F37, Stevenage Bioscience Catalyst, Gunnels Wood Road, Stevenage, Hertfordshire, SG1 2FX, UK}
\affil[3]{University of Oxford, Physical and Theoretical Chemistry Laboratory, Oxford, 610101, Oxfordshire, UK}
\affil[4]{IBM Research, IBM T.J. Watson Research Center, Yorktown Heights 10598, New York, USA}

\begin{document}

\maketitle

J.M. led the computational work and B.W. led the experimental work; these two authors contributed equally to this work \\

To whom correspondence should be addressed. E-mail: J.M. james.mcdonagh@serna.bio, B.W. bhwunsch@us.ibm.com

\section{\textbf{Abstract}}
The increasing importance of carbon capture technologies for deployment in remediating \coo{} emissions, and thus the necessity to improve capture materials to allow scalabiltiy and efficiency, faces the challenge of materials development, which can require substantial costs and time.  Machine learning offers a promising method for reducing the time and resource burdens of materials development through efficient correlation of structure-property relationships to allow down-selection and focusing on promising candidates.  Towards demonstrating this, we have developed an end-to-end "discovery cycle" to select new aqueous amines compatible with the commercially viable acid gas scrubbing carbon capture. We combine a simple, rapid laboratory assay for \coo{} absorption with a machine learning based molecular fingerprinting model approach.  The prediction process shows 60\% accuracy against experiment for both material parameters and 80\% for a single parameter on an external test set. The discovery cycle determined several promising amines that were verified experimentally, and which had not been applied to carbon capture previously.  In the process we have compiled a large, single-source data set for carbon capture amines and produced an open source machine learning tool for the identification of amine molecule candidates (https://github.com/IBM/Carbon-capture-fingerprint-generation).   

\section{\textbf{Introduction}}
Anthropogenic climate change will be a major world-wide concern of this century.\cite{wu2020solvent} Greenhouse gases such as \coo{}, methane and nitrous oxides (NO$_{x}$) are major contributing species to the climate emergency of global warming. Of these gases \coo{} is rightly receiving major attention, as it is the largest fraction of green house gases emitted and long lasting in the atmosphere.\cite{olivier2017trends} Recent reports\cite{Carrington2022Revealed} show that major oil and gas reserves continue to be exploited, with new fossil fuel burning infrastructure still being built.\cite{wu2020solvent} These activities add to the committed emissions from existing infrastructure that endanger climate targets.\cite{Tong2019} 

Carbon Capture, Utilization and Storage (CCUS) technologies, specifically those that target \coo{} emissions, have the potential for negative emissions and are likely to be necessary in meeting the Paris climate accord.\cite{wu2020solvent,bruhn2016separating} Although within CCUS there is growing interest in direct air capture projects to remove \coo{} from the air, this technology and its commercial deployment remains in its infancy, leaving point-source capture from post-combustion or heavy-emission industries the primary target for practical remediation \cite{Realmont2019}. CCUS technology can allow decarbonization of existing infrastructure and hard-to-abate emissions.\cite{iea2020} Recent publications have estimated there are 87 planned CCUS plants between 2020–2030\cite{iogp2020}.

Current CCUS technologies span carbon capture solvents, polymers or other membranes, and solid state materials. Of these technologies carbon capture solvents are the most mature, with on-going commercial usage and planned developments.\cite{carbonclean2023solvent, chao2020post, bui2018carbon} Carbon capture solvent technology is dominated by amine based solvents. Amines are common organic bases derived from ammonia by replacement of hydrogen with larger organic groups. Primary and secondary amines react with \coo{} through nucleophilic addition to form a carbamic acid (zwitterion intermediate) which is then further deprotenated by an additional amine (or base) to the carbmate; a process which leads to a 2:1 amine-to-\coo{} capture ratio (Figure \ref{scheme:prim_sec}). Tertiary amines cannot form the carbamic intermediate; instead they act on dissolved \coo{} as Bronsted bases to generate carbonates, allowing one amine to capture one \coo{} molecule (Figure \ref{scheme:tert}).\cite{said2020unified,puxty2009carbon, kenarsari2013review, yang2017computational} 

\begin{figure}[H]
\centering
\schemestart
 \chemname{\chemfig{HNR_1R_2}}{}
 \+{0pt,1em}
 \chemname{\chemfig{CO_2}}{}
 \arrow{<=>}
 \chemname{\chemfig{R_1R_2\chemabove{N}{\scriptstyle\oplus}HCO\chemabove{O}{\scriptstyle\ominus}}}{}
\schemestop
\label{scheme:prim_sec}
\end{figure}

\begin{figure}[H]
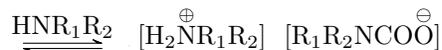

\centering
\schemestart
 \arrow{<=>[$\text{HNR}_{1}\text{R}_{2}$]}
  \chemname{ [\chemfig{H_{2}\chemabove{N}{\scriptstyle\oplus}R_1R_2}] }{}
  \chemname{[\chemfig{R_1R_2NCO\chemabove{O}{\scriptstyle\ominus}}]}{}
 \schemestop
 \caption{Primary and secondary amine general reaction scheme.}
 \label{scheme:prim_sec}
\end{figure}

\begin{figure}[H]
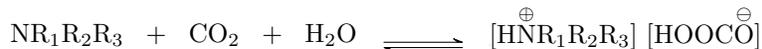

\centering
\schemestart
 \chemname{\chemfig{NR_1R_2R_3}}{}
 \+
 \chemname{\chemfig{CO_2}}{}
 \+
  \chemname{\chemfig{H_{2}O}}{}
 \arrow{<=>}
  \chemname{[\chemfig{H\chemabove{N}{\scriptstyle\oplus}R_1R_2R_3}]}{}
 \chemname{[\chemfig{HOOC\chemabove{O}{\scriptstyle\ominus}}]}{}
\schemestop
 \caption{Tertiary amine general reaction scheme.}
\label{scheme:tert}
\end{figure}

Post-combustion point-source capture is not widely adopted in any major carbon intensive industry (e.g. power generation, cement, heavy industry). Expanded adoption hinges on lower cost-per-ton-\coo{}, which can be improved through the amine solvent chemistry.  There is still a need for solvent development.\cite{Bernhardsen2017, raksajati2018comparison, raksajati2018solvent, orlov2022computational} Commonly used  amines for carbon capture include monoethanolamine (MEA), diethanolamine (DEA), piperazine (PZ), and methyldiethanolamine (MDEA), with state-of-the-art formulations often including these species. However, these solvents have drawbacks that include (but are not limited to) capture efficiency, thermal degradation, corrosion and, significantly, heat of regeneration.\cite{chao2020post} Substantial work has been made on screening and developing new carbon capture amine systems.\cite{orlov2022computational, mcdonagh2022chemical}. Experimental screening has provided diverse data sets of organic amines and alkaline solvents for carbon capture \cite{Puxty2009, Porcheron2011, Singh2007, Singh2009, Hussain2011, Conway2012, Conway2014, Chowdhury2013, Xiao2016, Yang2016, Bernhardsen2017, Kessler2018, Kessler2021}.  In addition to this, amines are also being explored as functionalization systems in solid state carbon capture for applications including DAC.\cite{hamdy2021application} Machine learning has been used to reconcile discrepancies in published carbon capture data to improve accuracy of modeling \cite{Suleman2017}.  In general, having a large data set analyzed with the same technique allows for more consistent cross-comparison of materials, enabling evaluation of structure-property relationships and down-selection for superior performance and further testing.

In this work, we detail a combined machine learning and  experimental loop for the discovery of several new promising amine candidate molecules for carbon capture, using an in-house generated data set of \coo{} absorption by various amine and organic nitrogen compounds.  The candidate molecules have been identified through computational screening followed by rapid laboratory testing. We detail here our screening methodology, predictive methods and experimental procedures. This process has also allowed us to create the largest set of single amine carbon capture performance metrics from a common experimental source available in the open literature, to the best of the authors' knowledge. This set comprises \ndatafull{} molecules. Some of the molecules in this data set overlap with other existing publications and find generally positive agreement in the property determinations. The majority of the data set however, appears to be novel. 

\section{Results}

\begin{figure}[H]
\centering
\includegraphics[width=\linewidth]{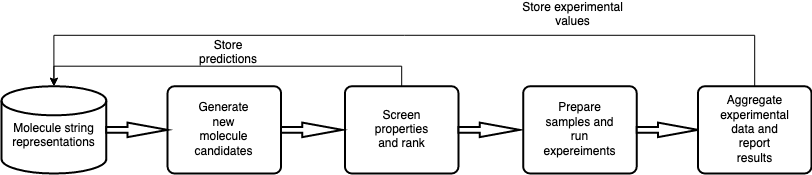}
\caption{Outline of molecular discovery cycle for carbon capture amines.}\label{fig:hl_methods}
\end{figure}

The process to survey amines for \coo{} capture reported here combines computational and experimental methods in an end-to-end discovery cycle and is shown schematically in Figure \ref{fig:hl_methods}.  We have previously developed a chemical fingerprinting technique that is successful at classifying and correlating molecules based on molecular fragments, showing improvement over other classification methods in the carbon capture space.\cite{mcdonagh2022chemical}  Based on this technique, we set out to develop the discovery cycle to train, validate and predict amine performance using in-house testing for all data generation and validation. 

\subsection{Validation of Classifier Models}\label{ss:val}

To establish the predictive capabilities of the system required first training and validating the machine learning models (classifiers). We used our fingerprinting technique to train binary classifiers on the molecule's performance based upon two key metrics of carbon capture: \textit{\ac{}} and \textit{\oir{}}.   The \ac{} is the molar ratio of \coo{} absorbed to amine moieties present (note: all capacities are \textit{per amine}, allowing comparison between mono- and polyamine molecules).  The \oir{} is the fitted pseudo-first order rate constant.  These metrics are central material properties for assessing how well a given amine captures \coo{}. The binary classifier outputs a 1 if a molecule's property is higher than a threshold and a 0 otherwise; thresholds are discussed in Methods and Materials. An initial training data set of \ntrain{} amines was generated using an in-house, rapid analysis instrument (see Methods and Materials) to assess the material properties.  A further validation set of \nval{} amines was used to assess the classifier performance.

We trained 10 classifiers for each metric and assessed their performance. Table \ref{tab:classifiers} presents the validation results for the top performing classifiers.  Results for all classifiers are given in \si{} section 4.  It is seen that for both \ac{} and \oir{} the classifiers achieve high accuracy ($>70\%$), with the \oir{} being the most accurately predicted of the two properties.  For \ac{} the sensitivity (rate of correct predictions for the positive class) and specificity (rate of correct predictions for the negative class) vary between the three top classifiers. For both \gp{} and \svm{} the classifiers are stronger in the prediction of the positive class, whilst for \ab{} it is stronger in the prediction of the negative class. Overall there is a marginally improved performance from \ab{} which is shown by the higher Receiver Operating Characteristic Area Under the Curve (ROC AUC) and Matthews Correction Coefficent (MCC) scores compared to the \gp{} and \svm{} models, and for this reason we selected the \ab{} model as our top capacity classifier.  For the \oir{} prediction we find that all three classifiers provide the same predictive performance to two decimal places, and have marginally improved metrics compared to the \ac{} metrics. We found from application of all three models to the external test set that the Gaussian Process model provides the best performance and was therefore selected as the presented model in this work. Results for the other models are provided in the \si{} section 4. Of note, all models show an accurate prediction of the negative class predicting all negative class molecules (5 of 11) to be in the negative class. The positive class prediction performance is less effective with the classifiers achieving 67\% correct positive class predictions.

\begin{table}[h!]
    \caption{Validation metrics from the op three models for predicting amine carbon capture performance. The full set of classifier performance metrics can be found in the \si{} section 4.  GP is Gaussian Process, AB is AdaBoost, SP is Support Vector Machine, DT is Decision Tree, XT is Extra Trees.  The training metrics are AC = \ac{}, OIR = \oir{}. ROC AUC is receiver operator curve, area under the curve, MCC is Matthews correlation constant.}
    \begin{tabular}{lrrrrr}
    \toprule
    Classifier &  Accuracy &  Sensitivity &  Specificity &  ROC AUC &       MCC \\
    \midrule
    GP\textunderscore{}AC &  0.73 &     0.83 &          0.60 & 0.72 &  0.45 \\
    AB\textunderscore{}AC &  0.73 &     0.67 &          0.80 & 0.73 &  0.47 \\
    SP\textunderscore{}AC &  0.73 &     0.83 &          0.60 & 0.72 &  0.45 \\
    GP\textunderscore{}OIR &  0.82 &     0.67 &          1.00 & 0.83 & 0.69 \\
    DT\textunderscore{}OIR &  0.82 &     0.67 &          1.00 & 0.83 & 0.69 \\
    XT\textunderscore{}OIR &  0.82 &     0.67 &          1.00 & 0.83 & 0.69 \\
    \bottomrule
    \end{tabular}
    \label{tab:classifiers}
\end{table}

\subsection{Prediction and Experimental Testing of Amine \coo{} Capture}

To assess the predictions of the amines, we used a validation set of \nval{} molecules to select the best performing models following training. These results are presented in table \ref{tab:classifiers} for the top three classifiers for each property. Having selected the top classifers, we next applied these to a test set of \ntest{} molecules. These results are shown in table \ref{tab:test_data}. These molecules were selected from our \ndatafull{} data set pseudo-randomly in that we required the sets to be approximately evenly split between the positive and negative classes for both properties. These compounds were tested experimentally with the rapid \coo{} absorption instrument and the data used to (1) compare the predictive performance of the classifiers and, (2) highlight molecules with high \ac{} and \oir{}.

\begin{figure*}[h]
\centering
\includegraphics[width=0.7\textwidth]{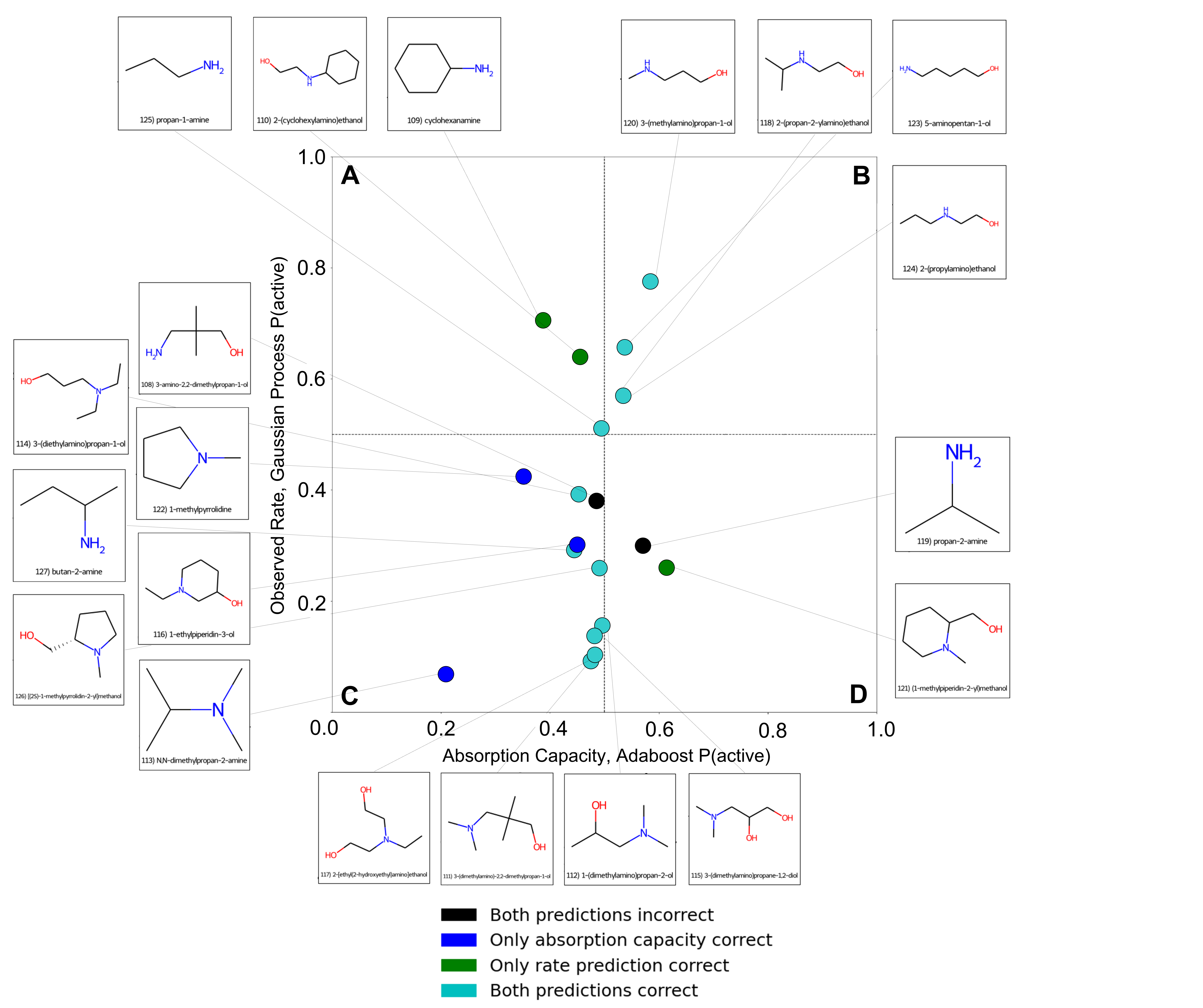}
\caption{Plots of the probability, from our classifier models. The \ac{} is predicted using the Adaboost classifier and \oir{} is predicted using the Gaussian Process. Panel A shows molecules with good probability for a promising \oir{} but unpromising \ac{}. Panel B shows molecules predicted well for both metrics. Panel C shows molecules with a low probability of either property being promising. Panel D molecules with good probability for a promising \ac{} but unpromising \oir{}.}\label{fig:computational_predictions_probability}
\end{figure*}

Figure \ref{fig:computational_predictions_probability} shows the  prediction results of \ac{} and \oir{}, and their assessment against the experimental values.  These axes are discretized based on the probability of a molecule being in the positive class, P(active). This gives a natural zero to one axes with 0.5 marking the boundary between the classes i.e. $< 0.5$ negative class and $\geq 0.5$ positive class.  The data are color coded based on if the predicted performance matches the experimental result. Panels A-D in Figure \ref{fig:computational_predictions_probability} show the chemical structures of the molecules predicted to be in each quadrant of the 2D plane.  Molecules in panel C are predicted to be unpromising for both \ac{} and \oir{} (panel c). Panels A and D show molecules predicted to be promising respectively for \oir{} or \ac{}, while panel B shows those molecules predicted to perform well for both metrics.  The prediction comparisons of the molecules for each quadrant, as well as the molecules in each of these categories based upon the experimental values of the properties, are shown in the \si{} Figures 4 and 5, respectively. Table \ref{tab:test_data} shows the predicted metrics for the displayed classifiers.

Considering the overall performance when combining these two classifiers we see that 60\% (4 of 7) of the molecules in the positive class for both properties are correctly predicted within the same quadrant as the experimental data. The remaining three molecules experimentally determined to be in this class are correctly predicted as promising for one of the two properties. We see a particularly strong performance in identifying the poor performing molecules; approximately 85\% (6 of 7) are found in the correct quadrant for both properties. In terms of screening for materials development this allows one to rapidly deprioritise molecules for testing with reasonable confidence. Of note is the promising performance obtained for these computationally efficient  statistical models with a limited training data set. We did not benchmark deep learning embeddings here largely due to the data set size for model fine tuning, and the context of this work being focused upon developing a discovery cycle. However, a recent benchmark has shown that molecular fingerprints often achieve better performance than deep learning embeddings for QSAR tasks such as the ones performed here \cite{Sabando2021UsingME}. The majority of top candidates predicted successfully are amino alcohols, either primary or secondary.  Although several amines have a beta hydroxyl, it seems the separation of the alcohol and amine groups is not a strong constraint given the examples of proply and pentyl derivatives. 

\begin{table}[h!]
    \tiny
    \centering
    \caption{Table of the predictions from the selected models for the external test set data. AC is \ac{} and OIR is \oir{}. Results for all models are given in the \si{}.}
    
    \label{tab:test_data}
    \begin{tabular}{llrrrrr}
    \toprule
    Property &                Model &  Accuracy &  Sensitivity &  Specificity &   ROC AUC &    MCC \\
    \midrule
    AC &        AdaBoost &  0.73 &     0.67 &          0.80 & 0.73 &  0.47 \\
     OIR & GaussianProcess &      0.75 &         0.58 &         1.00 &     0.79 & 0.60 \\
    \bottomrule
    \end{tabular}
\end{table}

\subsection{Data Set Exploration}

\begin{figure*}[h]
\centering
\includegraphics[width=0.7\textwidth]{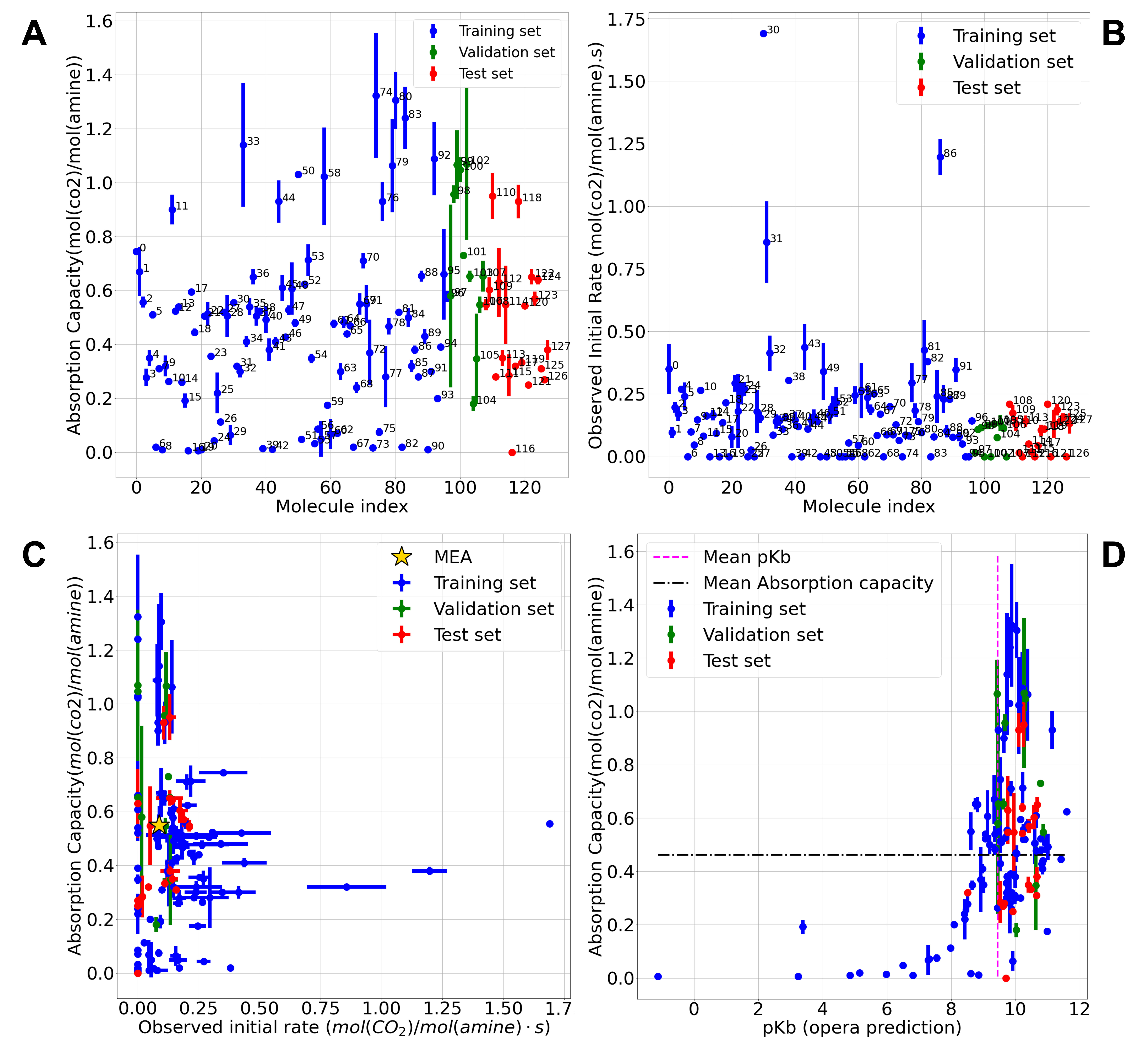}
\caption{Overview of the experimental data. Panel A: highlights the range of amine \ac{} explored in this data set. Panel B: Expounds upon the range of \oir{} explored in this data set. Panel C: Explores the relation between the two properties \ac{} and \oir{}; the ideal molecule would be positioned in the top right corner. P
anel D: Highlights the known trend of \ac{} with pKb which our data follows. Here we have predicted the pKb with the OPERA tools from the EPA.} \label{fig:experimental_dataset_graphs}
\end{figure*}

Figure \ref{fig:experimental_dataset_graphs} collects the \ndatafull{} experimentally measured  material properties for the amines used at the different computational stages of training the machine learning model (blue), validating (green) and testing  performance predictions (red). The molecules used in training and validation where selected as a sub-set from a wider computational screening and identification as outlined in sub-section \textcolor{red}{\ref{subsec:identification}}. The molecules in Figure \ref{fig:experimental_dataset_graphs} A and B are labelled by index from Table 1 in the \si{} which also gives the molecules IUPAC names and InChIkey's as identifiers. We provide all molecules SMILES, InChI, InChIkeys, IUPAC names and measure properties in the form of a csv file together with the fingerprint generation method at \codedataurl{} and a summary of the data used in the \si{} Table 1.  The data represents a diverse set of amines and organic nitrogen species, as seen in \si{} Figure 1.  Structurally, the data contains primary, secondary and tertiary amines, as well as polyamine combinations and other carbon-nitrogen moieties (\si{}, Figure 2).  Predominately the amines are aliphatic, but there is a small sub-set of aromatic compounds, all of which have primary amines.

Figure \ref{fig:experimental_dataset_graphs} (panels A and B) shows there is a wide range of values for the \ac{} and \oir{}, showing amines cover a wide parameter space in which to explore candidates.  Focusing on \ac{}, four clusters of values are observed (see \si{} Figure 3): species with \ac{} above 0.9 mol(\coo{})/mol(amine), between 0.45 and 0.7 mol(\coo{})/mol(amine), between 0.2 and 0.45 mol(\coo{})/mol(amine) and those below 0.2 mol(\coo{})/mol(amine). Based on the amine types, the two higher-value clusters correspond to tertiary/frustrated secondary amines (carbonate reaction mechanism) and primary/secondary amines (carbamate reaction mechanism).  The low \ac{} cluster represents poor or non-reaction species.  The 0.2 - 0.4 mol(\coo{})/mol(amine) is interesting in that it represents predominately primary and secondary amines that have reduced reactivity.

Panel C in Figure \ref{fig:experimental_dataset_graphs}  displays the concurrent behaviour of the molecules over both properties. It is clear in several cases that it is possible to achieve a high \ac{} but the \oir{} remains low. On this plane the ideal molecule would be at the top right corner. The molecule monoethanolamine (MEA) (a.k.a. 2-aminoethanol), a common standard for solvent based carbon capture studies, is highlighted using a star in Figure \ref{fig:experimental_dataset_graphs} C \cite{romeo2020comparative}. We see that in this space there are several molecules which improve on MEA for both properties concurrently. Finally, we turn to pKb. The pKa of the conjugate acid has been used as a indicator for carbon capture performance for many years. Here we have used the OPERA toolkit from the EPA to predict the molecule's basicity in terms of pKb. We see the expected trend of higher basicity correlating with higher capacity as has been seen previously \cite{Puxty2009}. 

\begin{figure}[H]
\centering
\includegraphics[width=\linewidth]{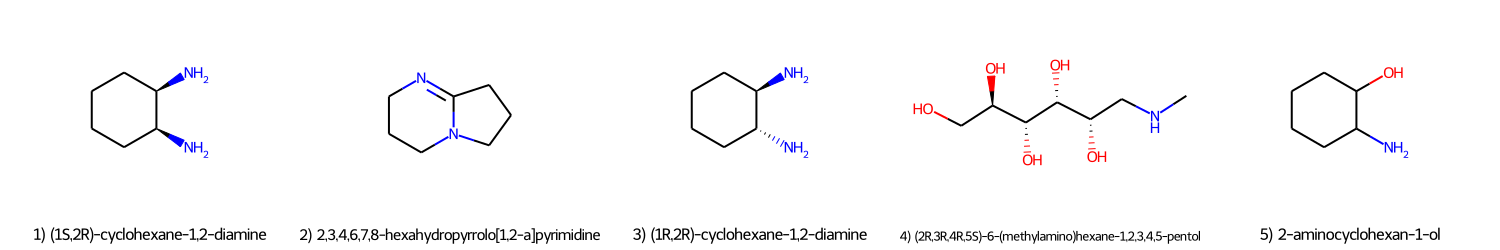}
\caption{Molecules 1 - 3 represent the highest reaction rate molecules we have encountered. Molecule 4 is a sugar derivative which shows promising carbon capture capabilities. Molecule 5 (which was tested as a commercially available racemic mixture) represents the best compromise between \ac{} and \oir{} in the current data set.} \label{fig:promising_molecules}
\end{figure}

Some of the most commonly used solvents, either independently or as part of a formulation, such as MEA and DEA \cite{romeo2020comparative} can also be used here as a benchmark. Doing this, we find 25 and 9 molecules in this data set which supersede both the \ac{} and \oir{} of these commonly used solvents, respectively (see \si{}). Of particular note within this set is (2R,3R,4R,5S)-6-(methylamino)hexane-1,2,3,4,5-pentol (N-methyl-D-glucamine) (Figure \ref{fig:promising_molecules}, image 4), which is a known additive for pharmaceuticals.  As an aminated sugar derivative, it suggests exploring other saccharides scaffolds for structure-function relationships in \coo{} capture. This molecule was also predicted to be the least toxic by the OPERA toolkit based on the CATMoS LD50 predictions. Potentially, such a molecule can provide a route to low carbon production of carbon capture agents. Of note also is 1-aminocyclohexan-1-ol (as a racemic mixture) which was found to have the highest relative \ac{} and \oir{} out of the data set, making it the optimal compromise molecule within in the data set.  That the MEA substructure (a well-known motif in carbon capture amines) is part of this compound, but with more constraint on its bond configurations due to the aliphatic cycle, it is of interest to consider if this preconditions the amine to a more favorable conformation for reacting with \coo{}; something a future study of all stereoisomers could resolve.

Finally, it is observed that, from the outset, the screening suggested some molecules with unusually high \oir{}. These three molecules are (1S,2R)-cyclohexane-1,2-diamine, (1R,2R)-cyclohexane-1,2-diamine and 2,3,4,6,7,8-hexahydropyrrolo[1,2-a]pyrimidine (1,5-diazabicyclo[4.3.0]non-5-ene, a.k.a. DBN) going from largest to smallest rate. Figure \ref{fig:promising_molecules}, images 1 - 3 shows the chemical structures of these molecules.  DBN is a well established organic base extensively used in synthesis, and is known to react with \coo{}.  That the diaminocyclohexane stereoisomers both show higher \oir{} compared to the majority of compounds could suggest a cooperatively effect due to the close proximity of the amines, though the differences in the stereochemistry may have a subtle effect on the coordination of \coo{} and thus the rate, given that the \textit{cis} isomer is moderately faster.

\section{Discussion}
New small molecule amines for carbon capture have been identified through a process of machine-learning driven computational screening coupled with laboratory scale evaluation of \ac{} and \oir{}. The results highlight the potential of data driven computational techniques in carbon capture chemistry. We have shown the use of a rapid-analysis instrument to assess \coo{} absorption in molecules and generate a sufficient data set for training successful, predictive models.  For a given test, only 200$\mu$L of sample is used, requiring 5-30 min, allowing a 100+ data set to be built up in a few weeks on a single instrument. The data set itself forms the largest single-source set of amines tested for carbon capture to date. 

We have shown that molecules in our training set, such as (2R,3R,4R,5S)-6-(methylamino)hexane-1,2,3,4,5-pentol, cyclohexane-1,2-diamine and 2-aminocyclohexan-1-ol have potential applications in carbon capture solvents. It is especially interesting to see molecules such as (2R,3R,4R,5S)-6-(methylamino)hexane-1,2,3,4,5-pentol which could be produced from waste biomass materials, as this is a cheap and likely low carbon feed stock material. Of note is the that cyclohexane-1,2-diamine  shows remarkable \oir{}. Although it exhibited an average \ac{} molecules such as this could be very useful in formulated blends to enhance the rate of \coo{} reactions.

The current work shows the value of data in the CCUS field. Using robust, widely deployable and computationally efficient models trained on fairly small data sets, has enabled promising screening of small amine based carbon capture solvents. This field remains relatively data poor, in part from the fact that much of the current formulation development has been done proprietorially for commercial use. The current work shows the potential value in opening \coo{} capture chemistry and formulation data and allowing computational modelling to be more extensively applied to these vital chemical technologies.  A centralized and curated source of carbon capture molecules could greatly improve machine learning and AI assisted development.

The structural space of organic amines is vast, and the results shown here reinforce the notion that there is still potential to locate new  molecules which can be utilized for carbon capture.  Of necessity for future work is to expand the training metrics to include material properties that become relevant at the industrial-scale deployment, including viscosity as a function of \coo{} loading, surface tension, corrosion, and oxidation and other degradation mechanisms, so that the large structure space can be probed more thoroughly.  Importantly is the inclusion of regeneration energy as an optimization parameter. In this work we have focused on using these molecules as independent, active capture systems i.e. a carbon capture solvent made up of amines diluted in water. However, there is scope to consider the utility of these molecules in formulated blends and in solid state materials such as MOFs and polymers, where these molecules may be able to improve upon the plain MOF or polymer skeleton.

The computational-experimental process and data set herein provide a baseline upon which further data can be generated and new models built and trialled. Ideally, this can form the basis for data standards and sharing to be discussed and applied within this field to enable the comparison of computational models. In particular, in this study we have chosen to not combine our dataset with existing ones to train machine learning models in order to use data from a single consistent experimental procedure. Further studies could investigate combining multiple data sources to expand on the screening models.

Carbon capture as a technology is advancing quickly. However, with many plants already under construction, more advanced and efficient carbon capture solvents are required urgently to help to mitigate and deliver negative emissions technology. Rapid experimental screening and computational modelling will enable researchers to advance this field at an enhanced rate in order to meet this challenge.

\section{Materials and Methods}

\subsection{Computational methods}
\begin{figure}[H]
\centering
\includegraphics[width=\linewidth]{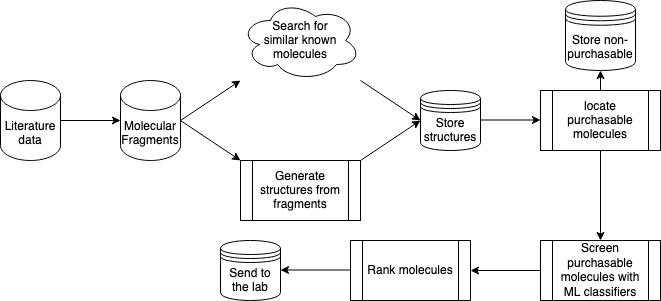}
\caption{Computational methods scheme outlining at a high level the process of molecule identification and high throughput virtual screening. }\label{fig:computer_meths}
\end{figure}

Figure \ref{fig:computer_meths} outlines the computational methods. The process is broken into two discrete stages: candidate molecule identification and high throughput screening of molecules. The authors have previously carried out detailed analysis of chemical representations and models for predicting \ac{} which inform the process used in the bottom row.\cite{mcdonagh2022chemical}

\subsubsection{Candidate Molecule Identification and Screening}\label{subsec:identification}
The identification stage is shown in the top row of Figure \ref{fig:computer_meths}. Here we apply tools more commonly associated with the pharmaceutical industry to identify molecules suitable for carbon capture. 

We have identified 164 unique amine molecules which have been reported in the literature\cite{puxty2009carbon, singh2007structure, singh2009structure, kim2015comparison, chowdhury2013co2, evjen2019aminoalkyl, hartono2017screening, rezaei2020molecular, yang2016toward} in relation to a range of carbon capture performance metrics such as absorption capacity, cyclic capacity and initial reaction rate. With this we created a data set by extracting string representations of these molecules from public databases PubChem\cite{kim2021pubchem} and ChemSpider\cite{pence2010chemspider}. We applied Matched Molecular Pairs (MMP) analysis to these molecules through MMPDB.\cite{dalke2018mmpdb} This analysis identifies molecular graph transformations up to three changes and generates transformation rules in the SMIRKS language.\cite{bone1999smiles} Transformations are rooted to an atom type based upon a molecular fingerprint which encodes the environment around an atom. Having identified these transformation rules and the atom type, the rules can then be applied to other molecules. The MMP method can then generate new molecular graphs by applying the rules to existing chemical graphs. This means that only atoms with very similar environments, as defined by the fingerprint, are considered equivalent and hence valid points of modification by applying the relevant SMIRKS. This allows us to deploy a data driven generative engine with only modest amounts of data related to chemical structures. This analysis was supplemented by molecular similarity searches on the PubChem data base using the generated structures.

Using this method we generated over 10,000 possible chemical structures. We filtered this set down in a step wise manner. Initially, we removed any duplicate molecules and any molecules with invalid string representations. We then predict important physical and chemical properties, such as water solubility, pKb and LD50 using the OPERA toolkit from the EPA\cite{mansouri2018opera} (please see \si{}). The current commercial offerings for carbon capture are predominately aqueous amines, hence water solubility is key. We additionally took note of the toxicity, via LD50, and pKb of the molecules from OPERA as highly toxic molecules are undesirable and very weak bases tend be poor for carbon capture. We also considered where possible the probability of a molecules possessing a promising \ac{} using model built on data from\cite{Puxty2009}. These factors were all considered and a manual down selection based upon these predictions and expert input was performed. This led to 287 molecules which formed our refined MMP set. Finally, we manually screened the compounds for (1) commercial availability at > 1 g scale and (2) compounds that represented unique structures/moieties. This later criterion was used to ensure a wider range of different organic amines (and organic nitrogen derivatives) was surveyed.  This down-selection yielded \ndatafull{} candidates.  

\subsubsection{High Throughput Screening}\label{subsec:hvts}
In our previous work we used 98 of these molecules to investigate classification model training and featurization \cite{mcdonagh2022chemical}. In the present work we extend the set of molecules and train new classifier models in over two properties \ac{} and \oir{}. We have trained our classifier models on a new random sample of \ntrain{} molecules, validated on \nval{} molecules ($\approx 90\%$ train $\approx 10\%$ validate) over both properties. The sets used are the same for both properties. We use CCS fingerprints as input features\cite{mcdonagh2022chemical} and apply Principle Component Analysis (PCA) to reduce the dimensionality, such that the PCA components explain 95\% of the variance. This leads to 21 components as features which we use to train our models to classify molecules. All models and training were carried out using scikit-learn version 1.0.2.\cite{scikit-learn}

Molecules were classified for each property separately. For \ac{} we followed the classification procedure defined in our previous work.\cite{mcdonagh2022chemical} This procedure classifies molecules based upon the amine types (primary, secondary, tertiary and poly) which make up the molecule and the likely reaction routes as defined in schemes \ref{scheme:prim_sec} and \ref{scheme:tert} for each amine group type. For the \oir{} we classified the molecules using MEA as a standard. We used our own measured value of MEA's \oir{} ($0.0868 \frac{\text{mol}_{\text{\coo{}}}}{\text{mol}_{\text{amine}}.\text{s}}$), which is in good agreement with the literature\cite{puxty2009carbon}. We therefore classified molecules whose \oir{} was $< 0.0868$ as the negative class and those whose \oir{} was $\geq 0.0868$ as the positive class.

We have applied these models to an external test set of \ntest{} molecules, which had not previously been measured to our knowledge, to fully test the screening capability. We have chosen these properties as they provide proxies for the thermodynamic capture (\ac{}) and the kinetic reaction rate (\oir{}). Considering a molecule in this 2D space allows us to seek the optimal balance between these often completing processes.  Given the limited data we have chosen a relatively small validation and test set to try to make the best use of the data in training our models, however this limits our ability to judge the generalizablity of the models following training. 

The total data set presented in this work is \ndatafull{} (only \ndata{} were used in model building as the data set contains some polymeric molecules which the current featureization is not suitable for). This set includes our previous set of molecules\cite{mcdonagh2022chemical} and adds new experimental data points. For each amine molecule we provide the measured properties of \ac{} and \oir{} in the \si{} Section 1. This is the largest single source data set for carbon capture amines in the open literature to the authors knowledge. 

We applied the CCS fingerprint representation \cite{mcdonagh2022chemical} as features for our models. We trained ten classifiers for each property and selected the best classifiers for each property using the validation set. 

The models were trained in a train test split fashion, performing a 10 fold cross validation to determine the optimal hyper-parameters from a limited grid search for each model over the training data (details of the hyper parameters are provided in the \si{} section 2). In each case the best parameters from this grid search were applied to each model and the validation set was applied to determine the models' performance.

A new unseen set of \ntest{} molecules was passed through the top performing classifiers and ranked based on the probability of each property being in the positive class.  These are the molecules tested by the classifiers and presented in the \textbf{Results} (Fig \ref{fig:computational_predictions_probability}). We additionally, supplemented these predictions with predictions of pKa and LD50 using the OPERA tools from the EPA.\cite{mansouri2018opera} These properties were not used for ranking, but provided further information on the molecules suitability for carbon capture. All predictions and properties are provided with the dataset (https://github.com/IBM/Carbon-capture-fingerprint-generation).

\subsection{Experimental methods}\label{sec:exp_method}
\begin{figure}[H]
\centering
\includegraphics[width=0.9\textwidth]{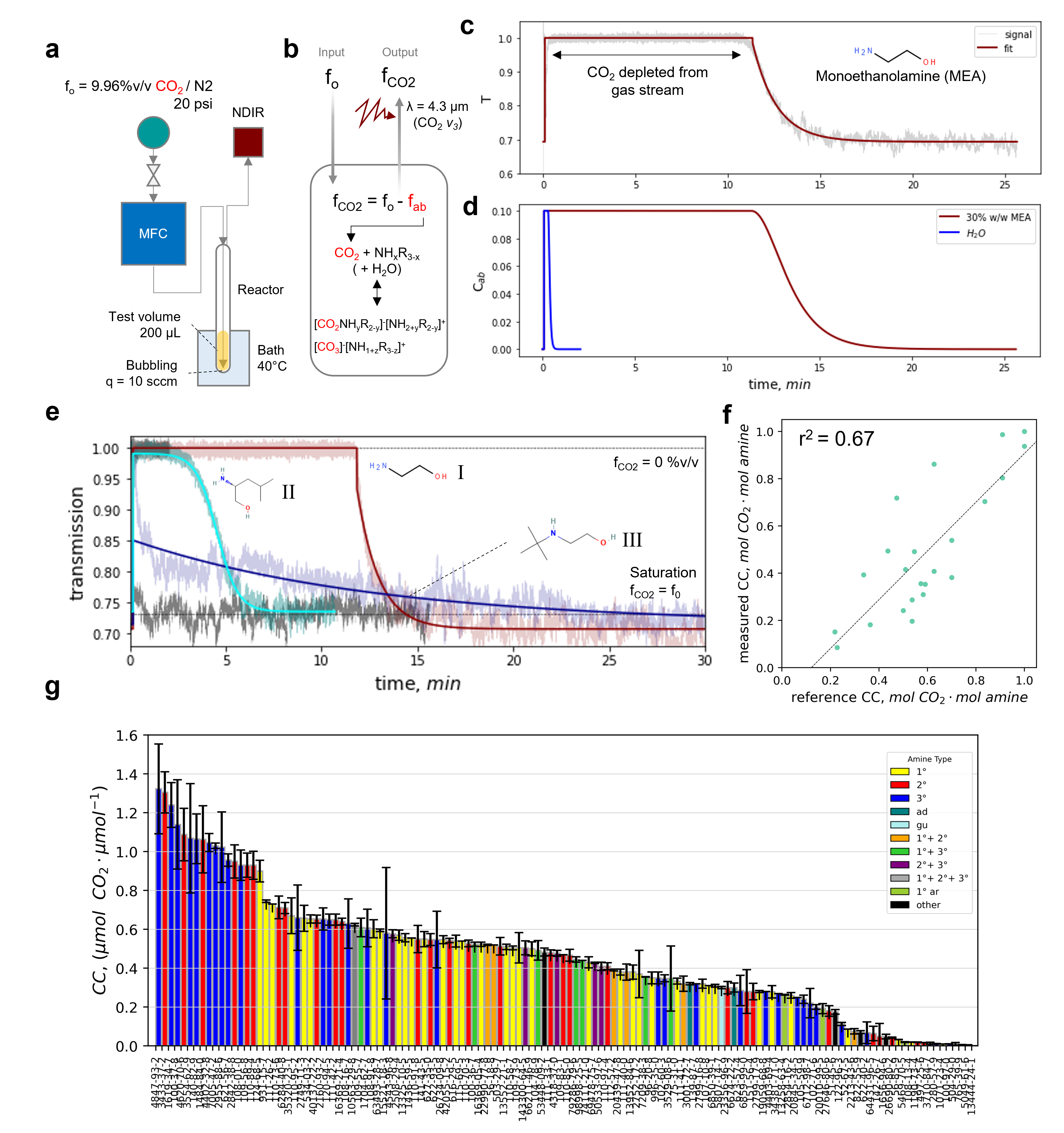}
\caption{Carbon dioxide absorption testing. \textbf{a}) Schematic representation of testing apparatus \textbf{b}) principle of analysis; gas is injected at a continuous rate and volume fraction $f_{o}$ of \coo{} into the aqueous amine solution.  The change in the \coo{} content of the exhaust stream is measured by infrared absorption at 4.3 $\mu$m.  \coo{} that reacts and absorbed by the solution is lost in the output gas stream, causing the measured fraction $f_{CO2}$ to be less than the supply fraction $f_{o}$. \textbf{c}) Representative signal of MEA absorption run, showing saturation region quantified by the breakthrough time, $t_{b}$ and roll-off to baseline ($f_{CO2}$ = $f_{o}$). \textbf{d}) Control signal of pure water absorption versus MEA. \textbf{e}) Absorption signals for I. MEA, II. L-leucinol and III. tert-butylaminoethanol.  Signals I and II exemplify fast reacting amines, showing either an exponential or sigmoidal type roll-off behavior (respectively).  \textbf{f}) Correlation of measured capture capacity $\alpha$ to that reported by Puxty \textit{et al} \cite{Puxty2009}.  \textbf{g}) Ranked capture capacity of tested samples labeled by amine type.
} \label{fig:expfig01}
\end{figure}

Aqueous amine solutions were tested for \coo{} absorption using a simple, in-house testing apparatus based on infrared absorbance (see Figure \ref{fig:expfig01}\textbf{a} and  \si{}). A gas stream of \coo{} / N$_{2}$ was bubbled into nominally 200 $\mu$L amine solution and the exhaust gas analyzed for \coo{} at the 4.3 $\mu$m absorption band (Figure \ref{fig:expfig01}\textbf{b}) A 3.9 $\mu$m reference band was used to account for slight attenuation due to humidity and signal drift.  The absorption signal was calibrated against 3 sources: atmosphere (taken to be 414 ppm), 9.96\%v/v \coo{} / balance nitrogen (hereafter referred to as 10\%v/v), and pure \coo{} as a function of flow rate, $q$ = 10 sccm.  Amine solutions were held at 40$\degree$C, chosen to fit typical industrial absorption operating temperature \cite{Notz2007}.  Signals were transformed from optical transmission into volume fraction \coo{} absorbed using a calibrated Modified Beer-Lambert equation (see \si{}). The measurement principle is that \coo{} lost in the exhaust stream must be absorbed in the amine solution; quantification of the gas content as a function of time and integration affords the total \coo{} absorbed and capture capacity, $\alpha$ (mol \coo{} $\cdot$ mol amine$^{-1}$).  

Monoethanolamine (MEA), 30\% w/w aqueous, was used as a calibrant as it has a well-established capture capacity of $\alpha$ = 0.50 mol CO2 / mol amine \cite{Puxty2009} (Figure \ref{fig:expfig01}\textbf{c}).  The estimated apparatus delay time is 0.16 min, and control experiments with pure water show a background absorption of $\sim$ 20 $\mu$mol \coo{} (Figure \ref{fig:expfig01}\textbf{d}). Analysis of the time progression of the absorbed \coo{} signal affords information on the relative speed of reaction (Figure \ref{fig:expfig01}\textbf{e}).

Benchmarking was performed with 23 reagents selected from the work of Puxty and co-workers, which had previously established a large dataset of 76 amines for carbon dioxide capture using a combination of microscale (100 $\mu$l) gravimetric analysis and macroscale (20 or 300 mL) liquid absorption analysis \cite{Puxty2009}.  The assay’s capture capacity shows a moderate correlation with the reference (Figure \ref{fig:expfig01}\textbf{f}) at $r^{2}$ = 0.67.  The current work recapitulates the capacity seen for fast reacting amines, as exemplified by MEA  ( $\alpha_{current}$ = 0.55 , $\alpha_{ref}$ = 0.49 / 0.56 ), as well as for slower tertiary / frustrated amines such as N,N-dimethylaminoethanol ($\alpha_{current}$ = 0.93 , $\alpha_{ref}$ = 0.92 ). 

130 aqueous amine solutions were prepared and tested for CO2 absorption (Figure \ref{fig:expfig01} $\textbf{g}$). Species include 1$\degree$, 2$\degree$ and 3$\degree$ amines, as well as polyamines, including both aliphatic and aromatic derivatives, and other organic nitrogen species (\si{} Figure 2 and 12). All solutions were prepared at 30$\%$ w/w (unless noted) as this concentration is typical in currently deployed industrial offerings.  Run times varied from 5 – 120 min to ensure a substantial saturated (i.e. no \coo{} absorbed) baseline was observed (analysis time can be shorted by 2x-4x if this criterion is removed).  Capture capacity (Figure \ref{fig:expfig01}\textbf{g}) shows the expected trend of $\alpha$ $\sim$ 1.0 for 3$\degree$ amines and sterically hindered 2$\degree$ / 1$\degree$ amines, and $\alpha$ $\sim$ 0.5 for 2$\degree$ / 1$\degree$ amines.  As seen in literature, there are middling cases where amine capacities are higher than expected, e.g. piperidine (2$\degree$) $\alpha$ = 0.93 ± 0.072, 2-amino-2-methyl-1-propanol (1$\degree$) $\alpha$ = 0.90 ± 0.056, and 2-aminocyclohexanol (1$\degree$) $\alpha$ = 0.75 ± 0.007, as well as a population of amines with poor performance, tailing to effectively non-reactive $\alpha$ = 0.006.  The distribution of capture capacity is multi-modal (\si{} Figure 6), with populations centered at $\alpha$ = 0.0, 0.27 and 0.49, as well as a low-sampled population at $\alpha$ $\sim$ 1.0.  The populations at 0.49 and $\sim$ 1.0 can be assigned to the expected carbamate and carbonate reaction paths.
\\
\\
\\
\textbf{Acknowledgement} The authors thank Mathias Steiner, Binquan Luan, James Hendrick and Nathaniel Park for insightful conversations. Data and required materials for this work can be found in the supporting information and at the following URL \codedataurl{}.

\bibliography{arXiv_2023newmolcCC_IBM_refs}

\begin{thebibliography}{10}

\bibitem{carbonclean2023solvent}
Carbon clean solutions limited: solvent-based-carbon-capture-refineries.
\newblock \url{}.
\newblock Accessed: 02-02-2023.

\bibitem{Bernhardsen2017}
Ida~M. Bernhardsen and Hanna~K. Knuutila.
\newblock A review of potential amine solvents for co2 absorption process:
  Absorption capacity, cyclic capacity and pka, 2017.

\bibitem{bone1999smiles}
Richard~GA Bone, Michael~A Firth, and Richard~A Sykes.
\newblock Smiles extensions for pattern matching and molecular transformations:
  Applications in chemoinformatics.
\newblock {\em Journal of chemical information and computer sciences},
  39(5):846--860, 1999.

\bibitem{bruhn2016separating}
Thomas Bruhn, Henriette Naims, and Barbara Olfe-Kr{\"a}utlein.
\newblock Separating the debate on co2 utilisation from carbon capture and
  storage.
\newblock {\em Environmental Science \& Policy}, 60:38--43, 2016.

\bibitem{bui2018carbon}
Mai Bui, Claire~S Adjiman, Andr{\'e} Bardow, Edward~J Anthony, Andy Boston,
  Solomon Brown, Paul~S Fennell, Sabine Fuss, Amparo Galindo, Leigh~A Hackett,
  et~al.
\newblock Carbon capture and storage (ccs): the way forward.
\newblock {\em Energy \& Environmental Science}, 11(5):1062--1176, 2018.

\bibitem{Carrington2022Revealed}
Damian Carrington and Matthew Taylor.
\newblock Revealed: the ‘carbon bombs’ set to trigger catastrophic climate
  breakdown.
\newblock {\em The Guardian}.

\bibitem{chao2020post}
Cong Chao, Yimin Deng, Raf Dewil, Jan Baeyens, and Xianfeng Fan.
\newblock Post-combustion carbon capture.
\newblock {\em Renewable and Sustainable Energy Reviews}, page 110490, 2020.

\bibitem{Chowdhury2013}
Firoz~Alam Chowdhury, Hidetaka Yamada, Takayuki Higashii, Kazuya Goto, and
  Masami Onoda.
\newblock Co2 capture by tertiary amine absorbents: A performance comparison
  study.
\newblock {\em Industrial and Engineering Chemistry Research}, 52:8323--8331, 6
  2013.

\bibitem{chowdhury2013co2}
Firoz~Alam Chowdhury, Hidetaka Yamada, Takayuki Higashii, Kazuya Goto, and
  Masami Onoda.
\newblock Co2 capture by tertiary amine absorbents: a performance comparison
  study.
\newblock {\em Industrial \& engineering chemistry research},
  52(24):8323--8331, 2013.

\bibitem{Conway2012}
William Conway, Xiaoguang Wang, Debra Fernandes, Robert Burns, Geoffrey
  Lawrance, Graeme Puxty, and Marcel Maeder.
\newblock Toward rational design of amine solutions for pcc applications: The
  kinetics of the reaction of co2(aq) with cyclic and secondary amines in
  aqueous solution.
\newblock {\em Environmental Science and Technology}, 46:7422--7429, 7 2012.

\bibitem{Conway2014}
William Conway, Qi~Yang, Susan James, Chiao~Chien Wei, Mark Bown, Paul Feron,
  and Graeme Puxty.
\newblock Designer amines for post combustion co2 capture processes.
\newblock volume~63, pages 1827--1834. Elsevier Ltd, 2014.

\bibitem{dalke2018mmpdb}
Andrew Dalke, Jerome Hert, and Christian Kramer.
\newblock mmpdb: An open-source matched molecular pair platform for large
  multiproperty data sets.
\newblock {\em Journal of chemical information and modeling}, 58(5):902--910,
  2018.

\bibitem{evjen2019aminoalkyl}
Sigvart Evjen, Oda~Siebke L{\o}ge, Anne Fiksdahl, and Hanna~K Knuutila.
\newblock Aminoalkyl-functionalized pyridines as high cyclic capacity co2
  absorbents.
\newblock {\em Energy \& Fuels}, 33(10):10011--10015, 2019.

\bibitem{hamdy2021application}
Louise~B Hamdy, Chitrakshi Goel, Jennifer~A Rudd, Andrew~R Barron, and Enrico
  Andreoli.
\newblock The application of amine-based materials for carbon capture and
  utilisation: an overarching view.
\newblock {\em Materials Advances}, 2021.

\bibitem{hartono2017screening}
Ardi Hartono, Solrun~Johanne Vevelstad, Arlinda Ciftja, and Hanna~K Knuutila.
\newblock Screening of strong bicarbonate forming solvents for co2 capture.
\newblock {\em International Journal of Greenhouse Gas Control}, 58:201--211,
  2017.

\bibitem{Hussain2011}
M.~Althaf Hussain, Yarasi Soujanya, and G.~Narahari Sastry.
\newblock Evaluating the efficacy of amino acids as co 2 capturing agents: A
  first principles investigation.
\newblock {\em Environmental Science and Technology}, 45:8582--8588, 10 2011.

\bibitem{iea2020}
International Energy~Association (IEA).
\newblock Ccus in clean energy transitions.
\newblock Technical report, International Energy Association, 2020.

\bibitem{kenarsari2013review}
Saeed~Danaei Kenarsari, Dali Yang, Guodong Jiang, Suojiang Zhang, Jianji Wang,
  Armistead~G Russell, Qiang Wei, and Maohong Fan.
\newblock Review of recent advances in carbon dioxide separation and capture.
\newblock {\em Rsc Advances}, 3(45):22739--22773, 2013.

\bibitem{Kessler2021}
Elmar Kessler, Luciana Ninni, Tanja Breug-Nissen, Benjamin Willy, Rolf
  Schneider, Muhammad Irfan, Jörn Rolker, Werner~R. Thiel, Erik von Harbou,
  and Hans Hasse.
\newblock Speciation in co2-loaded aqueous solutions of sixteen
  triacetoneamine-derivates (evas) and elucidation of structure-property
  relationships.
\newblock {\em Chemical Engineering Science}, 229, 1 2021.

\bibitem{Kessler2018}
Elmar Kessler, Luciana~Ninni Schäfer, Benjamin Willy, Rolf Schneider, Muhammad
  Irfan, Jörn Rolker, Erik~Von Harbou, and Hans Hasse.
\newblock Structure-property relationships for new amines for reactive co2
  absorption.
\newblock {\em Chemical Engineering Transactions}, 69:109--114, 2018.

\bibitem{kim2021pubchem}
Sunghwan Kim, Jie Chen, Tiejun Cheng, Asta Gindulyte, Jia He, Siqian He,
  Qingliang Li, Benjamin~A Shoemaker, Paul~A Thiessen, Bo~Yu, et~al.
\newblock Pubchem in 2021: new data content and improved web interfaces.
\newblock {\em Nucleic acids research}, 49(D1):D1388--D1395, 2021.

\bibitem{kim2015comparison}
Young~Eun Kim, Soung~Hee Yun, Jeong~Ho Choi, Sung~Chan Nam, Sung~Youl Park,
  Soon~Kwan Jeong, and Yeo~Il Yoon.
\newblock Comparison of the co2 absorption characteristics of aqueous solutions
  of diamines: absorption capacity, specific heat capacity, and heat of
  absorption.
\newblock {\em Energy \& Fuels}, 29(4):2582--2590, 2015.

\bibitem{mansouri2018opera}
Kamel Mansouri, Chris~M Grulke, Richard~S Judson, and Antony~J Williams.
\newblock Opera models for predicting physicochemical properties and
  environmental fate endpoints.
\newblock {\em Journal of cheminformatics}, 10(1):1--19, 2018.

\bibitem{mcdonagh2022chemical}
James McDonagh, Stamatia Zavitsanou, Alexander Harrison, Dimitry Zubarev,
  Benjamin Wunsch, Theordore van Kessel, and Flaviu Cipcigan.
\newblock Chemical space analysis and property prediction for carbon capture
  amine molecules.
\newblock 2022.

\bibitem{Notz2007}
R.~Notz, N.~Asprion, I.~Clausen, and H.~Hasse.
\newblock Selection and pilot plant tests of new absorbents for post-combustion
  carbon dioxide capture.
\newblock {\em Chemical Engineering Research and Design}, 85:510--515, 2007.

\bibitem{iogp2020}
International~Association of~Oil and Gas Producers.
\newblock Map of global ccus projects, 2020.

\bibitem{olivier2017trends}
Jos~GJ Olivier, KM~Schure, and JAHW Peters.
\newblock Trends in global co2 and total greenhouse gas emissions.
\newblock {\em PBL Netherlands Environmental Assessment Agency}, 5, 2017.

\bibitem{orlov2022computational}
Alexey~A Orlov, Alain Valtz, Christophe Coquelet, Xavier Rozanska, Erich
  Wimmer, Gilles Marcou, Dragos Horvath, B{\'e}n{\'e}dicte Poulain, Alexandre
  Varnek, and Fr{\'e}d{\'e}rick de~Meyer.
\newblock Computational screening methodology identifies effective solvents for
  co2 capture.
\newblock {\em Communications Chemistry}, 5(1):1--7, 2022.

\bibitem{scikit-learn}
F.~Pedregosa, G.~Varoquaux, A.~Gramfort, V.~Michel, B.~Thirion, O.~Grisel,
  M.~Blondel, P.~Prettenhofer, R.~Weiss, V.~Dubourg, J.~Vanderplas, A.~Passos,
  D.~Cournapeau, M.~Brucher, M.~Perrot, and E.~Duchesnay.
\newblock Scikit-learn: Machine learning in {P}ython.
\newblock {\em Journal of Machine Learning Research}, 12:2825--2830, 2011.

\bibitem{pence2010chemspider}
Harry~E Pence and Antony Williams.
\newblock Chemspider: an online chemical information resource, 2010.

\bibitem{Porcheron2011}
Fabien Porcheron, Alexandre Gibert, Pascal Mougin, and Aurélie Wender.
\newblock High throughput screening of co 2 solubility in aqueous monoamine
  solutions.
\newblock {\em Environmental Science and Technology}, 45:2486--2492, 3 2011.

\bibitem{puxty2009carbon}
Graeme Puxty, Robert Rowland, Andrew Allport, Qi~Yang, Mark Bown, Robert Burns,
  Marcel Maeder, and Moetaz Attalla.
\newblock Carbon dioxide postcombustion capture: a novel screening study of the
  carbon dioxide absorption performance of 76 amines.
\newblock {\em Environmental science \& technology}, 43(16):6427--6433, 2009.

\bibitem{Puxty2009}
Graeme Puxty, Robert Rowland, Andrew Allport, Qi~Yang, Mark Bown, Robert Burns,
  Marcel Maeder, and Moetaz Attalla.
\newblock Carbon dioxide postcombustion capture: A novel screening study of the
  carbon dioxide absorption performance of 76 amines.
\newblock {\em Environmental Science and Technology}, 43:6427--6433, 8 2009.

\bibitem{raksajati2018solvent}
Anggit Raksajati, Minh Ho, and Dianne Wiley.
\newblock Solvent development for post-combustion co2 capture: Recent
  development and opportunities.
\newblock In {\em MATEC Web of Conferences}, volume 156, page 03015. EDP
  Sciences, 2018.

\bibitem{raksajati2018comparison}
Anggit Raksajati, Minh~T Ho, and Dianne~E Wiley.
\newblock Comparison of solvent development options for capture of co2 from
  flue gases.
\newblock {\em Industrial \& Engineering Chemistry Research},
  57(19):6746--6758, 2018.

\bibitem{Realmont2019}
G.~Realmonte, L.~Drouet, A.~Gambhir, J.~Glynn, A.~Hawkes, A.~C. Köberle, and
  M.~Tavoni.
\newblock {A}n inter-model assessment of the role of direct air capture in deep
  mitigation pathways.
\newblock {\em {N}at. {C}ommun.}, 10:3277, 2019.

\bibitem{rezaei2020molecular}
Bijan Rezaei, Siavash Riahi, and Ali~Ebrahimpoor Gorji.
\newblock Molecular investigation of amine performance in the carbon capture
  process: least squares support vector machine approach.
\newblock {\em Korean Journal of Chemical Engineering}, 37(1):72--79, 2020.

\bibitem{romeo2020comparative}
Luis~M Romeo, Diego Minguell, Reza Shirmohammadi, and Jose{\'{e}}~M
  Andr{\'{e}}s.
\newblock Comparative analysis of the efficiency penalty in power plants of
  different amine-based solvents for co2 capture.
\newblock {\em Industrial \& Engineering Chemistry Research},
  59(21):10082--10092, 2020.

\bibitem{Sabando2021UsingME}
Mar{\'i}a~Virginia Sabando, Ignacio Ponzoni, Evangelos~E. Milios, and Axel~J.
  Soto.
\newblock Using molecular embeddings in qsar modeling: Does it make a
  difference?
\newblock {\em Briefings in bioinformatics}, 2021.

\bibitem{said2020unified}
Ridha~Ben Said, Joel~Motaka Kolle, Khaled Essalah, Bahoueddine Tangour, and
  Abdelhamid Sayari.
\newblock A unified approach to co2--amine reaction mechanisms.
\newblock {\em ACS omega}, 5(40):26125--26133, 2020.

\bibitem{Singh2007}
Prachi Singh, John~P.M. Niederer, and Geert~F. Versteeg.
\newblock Structure and activity relationships for amine based co2
  absorbents-i.
\newblock {\em International Journal of Greenhouse Gas Control}, 1:5--10, 2007.

\bibitem{singh2007structure}
Prachi Singh, John~PM Niederer, and Geert~F Versteeg.
\newblock Structure and activity relationships for amine based co2
  absorbents—i.
\newblock {\em International Journal of Greenhouse Gas Control}, 1(1):5--10,
  2007.

\bibitem{Singh2009}
Prachi Singh, John~P.M. Niederer, and Geert~F. Versteeg.
\newblock Structure and activity relationships for amine-based co2
  absorbents-ii.
\newblock {\em Chemical Engineering Research and Design}, 87:135--144, 2009.

\bibitem{singh2009structure}
Prachi Singh, John~PM Niederer, and Geert~F Versteeg.
\newblock Structure and activity relationships for amine-based co2
  absorbents-ii.
\newblock {\em Chemical Engineering Research and Design}, 87(2):135--144, 2009.

\bibitem{Suleman2017}
Humbul Suleman, Abdulhalim~Shah Maulud, and Zakaria Man.
\newblock Reconciliation of outliers in co2-alkanolamine-h2o datasets by robust
  neural network winsorization.
\newblock {\em Neural Computing and Applications}, 28:2621--2632, 9 2017.

\bibitem{Tong2019}
Dan Tong, Qiang Zhang, Yixuan Zheng, Ken Caldeira, Christine Shearer, Chaopeng
  Hong, Yue Qin, and Steven~J. Davis.
\newblock Committed emissions from existing energy infrastructure jeopardize
  1.5{\hspace{0.167em}}{\textdegree}c climate target.
\newblock {\em Nature}, 572(7769):373--377, July 2019.

\bibitem{wu2020solvent}
Xiao Wu, Meihong Wang, Peizhi Liao, Jiong Shen, and Yiguo Li.
\newblock Solvent-based post-combustion co2 capture for power plants: A
  critical review and perspective on dynamic modelling, system identification,
  process control and flexible operation.
\newblock {\em Applied Energy}, 257:113941, 2020.

\bibitem{Xiao2016}
Min Xiao, Helei Liu, Raphael Idem, Paitoon Tontiwachwuthikul, and Zhiwu Liang.
\newblock A study of structure–activity relationships of commercial tertiary
  amines for post-combustion co2 capture.
\newblock {\em Applied Energy}, 184:219--229, 12 2016.

\bibitem{Yang2016}
Qi~Yang, Graeme Puxty, Susan James, Mark Bown, Paul Feron, and William Conway.
\newblock Toward intelligent co2 capture solvent design through experimental
  solvent development and amine synthesis.
\newblock {\em Energy and Fuels}, 30:7503--7510, 9 2016.

\bibitem{yang2016toward}
Qi~Yang, Graeme Puxty, Susan James, Mark Bown, Paul Feron, and William Conway.
\newblock Toward intelligent co2 capture solvent design through experimental
  solvent development and amine synthesis.
\newblock {\em Energy \& Fuels}, 30(9):7503--7510, 2016.

\bibitem{yang2017computational}
Xin Yang, Robert~J Rees, William Conway, Graeme Puxty, Qi~Yang, and David~A
  Winkler.
\newblock Computational modeling and simulation of co2 capture by aqueous
  amines.
\newblock {\em Chemical reviews}, 117(14):9524--9593, 2017.

\end{thebibliography}
\bibliographystyle{plain}

\end{document}


\maketitle

\begin{landscape}
\section{Data set}
\tiny
\begin{longtable}{lllrrrr}
\caption{Molecules CAS Number, InChI key , Systematic IUPAC name, \ac{} and \ac{} standard deviation in units of mol(\coo{})/mol(amine), \oir{} and \oir{} standard deviation in units of mol(\coo{})/mol(amine).s for all molecules in the training, testing and validation data set.}\\
\toprule
CAS Number & InChI keys & IUPAC names & AC & AC $\sigma$ & OIR & OIR $\sigma$\\
\midrule
\endfirsthead
\caption[]{Raw data} \\
\toprule
CAS Number & InChI keys & IUPAC names & AC & AC $\sigma$ & OIR & OIR $\sigma$\\
\midrule
\endhead
\midrule
\multicolumn{3}{r}{{Continued on next page}} \\
\midrule
\endfoot

\bottomrule
\endlastfoot

   931-15-7 &   PQMCFTMVQORYJC-UHFFFAOYSA-N &                             2-aminocyclohexan-1-ol &      0.74 &         0.01 & 0.35 &  0.10 \\
 35320-23-1 &   BKMMTJMQCTUHRP-GSVOUGTGSA-N &                            (2R)-2-aminopropan-1-ol &      0.67 &         0.09 & 0.10 &  0.02 \\
 13325-10-5 &   BLFRQYKZFKYQLO-UHFFFAOYSA-N &                                  4-aminobutan-1-ol &      0.56 &         0.02 & 0.20 &  0.02 \\
  4403-69-4 &   GVOYKJPMUUJXBS-UHFFFAOYSA-N &                             2-(aminomethyl)aniline &      0.28 &         0.03 & 0.17 &  0.03 \\
    96-20-8 & JCBPETKZIGVZRE-UHFFFAOYSA-N   &                                  2-aminobutan-1-ol &      0.35 &         0.03 & 0.27 &  0.01 \\
 13552-21-1 &   KODLUXHSIZOKTG-UHFFFAOYSA-N &                                  1-aminobutan-2-ol &      0.51 &         0.01 & 0.24 &  0.06 \\
  1190-73-4 &   AXFZADXWLMXITO-UHFFFAOYSA-N &                       N-(2-sulfanylethyl)acetamide &      0.02 &         0.00 & 0.00 &  0.00 \\
  6850-39-1 & NIQIPYGXPZUDDP-UHFFFAOYSA-N   &                             3-aminocyclohexan-1-ol &      0.31 &         0.01 & 0.10 &  0.00 \\
  1071-23-4 &   SUHOOTKUPISOBE-UHFFFAOYSA-N &                  2-aminoethyl dihydrogen phosphate &      0.01 &         0.00 & 0.05 &  0.00 \\
   111-41-1 & LHIJANUOQQMGNT-UHFFFAOYSA-N   &                       2-(2-aminoethylamino)ethanol &      0.32 &         0.04 & 0.15 &  0.01 \\
 13258-63-4 &   IDLHTECVNDEOIY-UHFFFAOYSA-N &                           2-pyridin-4-ylethanamine &      0.26 &         0.00 & 0.27 &  0.01 \\
   124-68-5 &   CBTVGIZVANVGBH-UHFFFAOYSA-N &                        2-amino-2-methylpropan-1-ol &      0.90 &         0.06 & 0.08 &  0.01 \\
 16369-05-4 &   NWYYWIJOWOLJNR-UHFFFAOYSA-N &                         2-amino-3-methylbutan-1-ol &      0.52 &         0.01 & 0.16 &  0.01 \\
   115-69-5 & UXFQFBNBSPQBJW-UHFFFAOYSA-N   &                   2-amino-2-methylpropane-1,3-diol &      0.54 &         0.00 & 0.00 &  0.00 \\
  2854-16-2 & LXQMHOKEXZETKB-UHFFFAOYSA-N   &                        1-amino-2-methylpropan-2-ol &      0.26 &         0.01 & 0.17 &  0.02 \\
 20010-99-5 &   HQIBSDCOMQYSPF-UHFFFAOYSA-N &                            pyrazin-2-ylmethanamine &      0.19 &         0.03 & 0.09 &  0.01 \\
  5049-61-6 &   XFTQRUTUGRCSGO-UHFFFAOYSA-N &                                    pyrazin-2-amine &      0.01 &         0.00 & 0.00 &  0.00 \\
 63493-28-7 &   IGEIPFLJVCPEKU-UHFFFAOYSA-N &                                     pentan-2-amine &      0.59 &         0.01 & 0.14 &  0.01 \\
 79286-79-6 &   NGXSWUFDCSEIOO-UHFFFAOYSA-N &                                 pyrrolidin-3-amine &      0.45 &         0.01 & 0.21 &  0.01 \\
   765-39-9 &   YNZAFFFENDLJQG-UHFFFAOYSA-N &                                     pyrrol-1-amine &      0.01 &         0.00 & 0.00 &  0.00 \\
   504-29-0 & ICSNLGPSRYBMBD-UHFFFAOYSA-N   &                                    pyridin-2-amine &      0.01 &         0.01 & 0.08 &  0.04 \\
   110-58-7 &   DPBLXKKOBLCELK-UHFFFAOYSA-N &                                     pentan-1-amine &      0.51 &         0.01 & 0.29 &  0.03 \\
   503-29-7 &   HONIICLYMWZJFZ-UHFFFAOYSA-N &                                          azetidine &      0.51 &         0.05 & 0.18 &  0.15 \\
  7209-38-3 & XUSNPFGLKGCWGN-UHFFFAOYSA-N   &  3-[4-(3-aminopropyl)piperazin-1-yl]propan-1-amine &      0.36 &         0.00 & 0.25 &  0.05 \\
 26690-80-2 &   GPTXCAZYUMDUMN-UHFFFAOYSA-N &             tert-butyl N-(2-hydroxyethyl)carbamate &      0.04 &         0.00 & 0.27 &  0.03 \\
  6712-98-7 &   ZFECCYLNALETDE-UHFFFAOYSA-N &            1-[bis(2-hydroxyethyl)amino]propan-2-ol &      0.22 &         0.08 & 0.00 &  0.00 \\
   122-96-3 &   VARKIGWTYBUWNT-UHFFFAOYSA-N &        2-[4-(2-hydroxyethyl)piperazin-1-yl]ethanol &      0.11 &         0.01 & 0.03 &  0.00 \\
    56-18-8 &   OTBHHUPVCYLGQO-UHFFFAOYSA-N &              N'-(3-aminopropyl)propane-1,3-diamine &      0.52 &         0.00 & 0.00 &  0.00 \\
   109-73-9 &   HQABUPZFAYXKJW-UHFFFAOYSA-N &                                      butan-1-amine &      0.51 &         0.08 & 0.18 &  0.08 \\
 64431-96-5 &   HHKZCCWKTZRCCL-UHFFFAOYSA-N & 2-[3-[[1,3-dihydroxy-2-(hydroxymethyl)propan-2-... &      0.06 &         0.04 & 0.15 &  0.02 \\
  1436-59-5 & SSJXIUAHEKJCMH-OLQVQODUSA-N   &                    (1S,2R)-cyclohexane-1,2-diamine &      0.56 &         0.00 & 1.69 &  0.00 \\
  3001-72-7 & SGUVLZREKBPKCE-UHFFFAOYSA-N   &      2,3,4,6,7,8-hexahydropyrrolo[1,2-a]pyrimidine &      0.32 &         0.01 & 0.86 &  0.16 \\
  6674-22-2 & GQHTUMJGOHRCHB-UHFFFAOYSA-N   &   2,3,4,6,7,8,9,10-octahydropyrimido[1,2-a]azepine &      0.30 &         0.02 & 0.41 &  0.07 \\
   100-37-8 & BFSVOASYOCHEOV-UHFFFAOYSA-N   &                            2-(diethylamino)ethanol &      1.14 &         0.23 & 0.09 &  0.01 \\
 50533-97-6 &   YFJAIURZMRJPDB-UHFFFAOYSA-N &                      N,N-dimethylpiperidin-4-amine &      0.41 &         0.02 & 0.14 &  0.03 \\
   616-29-5 & UYBWIEGTWASWSR-UHFFFAOYSA-N   &                             1,3-diaminopropan-2-ol &      0.54 &         0.03 & 0.14 &  0.02 \\
   111-42-2 &   ZBCBWPMODOFKDW-UHFFFAOYSA-N &                     2-(2-hydroxyethylamino)ethanol &      0.65 &         0.03 & 0.11 &  0.00 \\
143300-64-5 &   OUXRMEUJNPVXMM-UHFFFAOYSA-N &                       N,N-diethylpiperidin-4-amine &      0.51 &         0.04 & 0.15 &  0.01 \\
   100-36-7 &   UDGSVBYJWHOHNN-UHFFFAOYSA-N &                    N',N'-diethylethane-1,2-diamine &      0.52 &         0.01 & 0.30 &  0.01 \\
  3710-84-7 &   FVCOIAYSJZGECG-UHFFFAOYSA-N &                           N,N-diethylhydroxylamine &      0.01 &         0.01 & 0.00 &  0.00 \\
   104-78-9 &   QOHMWDJIBGVPIF-UHFFFAOYSA-N &                   N',N'-diethylpropane-1,3-diamine &      0.49 &         0.05 & 0.15 &  0.03 \\
   111-40-0 & RPNUMPOLZDHAAY-UHFFFAOYSA-N   &                N'-(2-aminoethyl)ethane-1,2-diamine &      0.38 &         0.04 & 0.12 &  0.01 \\
   280-57-9 &   IMNIMPAHZVJRPE-UHFFFAOYSA-N &                      1,4-diazabicyclo[2.2.2]octane &      0.01 &         0.00 & 0.00 &  0.00 \\
   110-97-4 & LVTYICIALWPMFW-UHFFFAOYSA-N   &                1-(2-hydroxypropylamino)propan-2-ol &      0.41 &         0.02 & 0.44 &  0.09 \\
   108-01-0 & UEEJHVSXFDXPFK-UHFFFAOYSA-N   &                           2-(dimethylamino)ethanol &      0.93 &         0.08 & 0.11 &  0.00 \\
   109-55-7 & IUNMPGNGSSIWFP-UHFFFAOYSA-N   &                  N',N'-dimethylpropane-1,3-diamine &      0.61 &         0.05 & 0.15 &  0.02 \\
 69478-75-7 &   AVAWMINJNRAQFS-UHFFFAOYSA-N &                     N,N-dimethylpyrrolidin-3-amine &      0.43 &         0.00 & 0.16 &  0.03 \\
   110-70-3 & KVKFRMCSXWQSNT-UHFFFAOYSA-N   &                    N,N'-dimethylethane-1,2-diamine &      0.53 &         0.02 & 0.14 &  0.01 \\
  1704-62-7 &   YSAANLSYLSUVHB-UHFFFAOYSA-N &                 2-[2-(dimethylamino)ethoxy]ethanol &      0.61 &         0.10 & 0.00 &  0.00 \\
   111-33-1 &   UQUPIHHYKUEXQD-UHFFFAOYSA-N &                   N,N'-dimethylpropane-1,3-diamine &      0.48 &         0.01 & 0.34 &  0.11 \\
  7005-47-2 &   XRIBIDPMFSLGFS-UHFFFAOYSA-N &              2-(dimethylamino)-2-methylpropan-1-ol &      1.03 &         0.01 & 0.00 &  0.00 \\
  1656-94-6 &   FEBQXMFOLRVSGC-UHFFFAOYSA-N &   1-pyridin-3-yl-N-(pyridin-3-ylmethyl)methanamine &      0.05 &         0.01 & 0.17 &  0.04 \\
 10563-29-8 &   OMKZWUPRGQMQJC-UHFFFAOYSA-N &    N'-[3-(dimethylamino)propyl]propane-1,3-diamine &      0.62 &         0.01 & 0.20 &  0.04 \\
   110-73-6 &   MIJDSYMOBYNHOT-UHFFFAOYSA-N &                              2-(ethylamino)ethanol &      0.71 &         0.06 & 0.22 &  0.06 \\
   102-60-3 & NSOXQYCFHDMMGV-UHFFFAOYSA-N   & 1-[2-[bis(2-hydroxypropyl)amino]ethyl-(2-hydrox... &      0.35 &         0.02 & 0.00 &  0.00 \\
    56-85-9 &   ZDXPYRJPNDTMRX-VKHMYHEASA-N &               (2S)-2,5-diamino-5-oxopentanoic acid &      0.03 &         0.00 & 0.00 &  0.00 \\
    52-67-5 &   VVNCNSJFMMFHPL-VKHMYHEASA-N &      (2S)-2-amino-3-methyl-3-sulfanylbutanoic acid &      0.09 &         0.00 & 0.00 &  0.00 \\
   142-26-7 &   PVCJKHHOXFKFRP-UHFFFAOYSA-N &                        N-(2-hydroxyethyl)acetamide &      0.05 &         0.07 & 0.05 &  0.00 \\
  2955-88-6 &   XBRDBODLCHKXHI-UHFFFAOYSA-N &                           2-pyrrolidin-1-ylethanol &      1.02 &         0.18 & 0.00 &  0.00 \\
    51-45-6 &   NTYJJOPFIAHURM-UHFFFAOYSA-N &                     2-(1H-imidazol-5-yl)ethanamine &      0.17 &         0.01 & 0.24 &  0.04 \\
   622-40-2 &   KKFDCBRMNNSAAW-UHFFFAOYSA-N &                            2-morpholin-4-ylethanol &      0.07 &         0.06 & 0.04 &  0.00 \\
  4318-37-0 &   FXHRAKUEZPSMLJ-UHFFFAOYSA-N &                             1-methyl-1,4-diazepane &      0.48 &         0.02 & 0.26 &  0.11 \\
   822-55-9 &   QDYTUZCWBJRHKK-UHFFFAOYSA-N &                           1H-imidazol-5-ylmethanol &      0.07 &         0.01 & 0.00 &  0.00 \\
 23356-96-9 & HVVNJUAVDAZWCB-YFKPBYRVSA-N   &                     [(2S)-pyrrolidin-2-yl]methanol &      0.30 &         0.03 & 0.24 &  0.05 \\
 53448-09-2 &   VPSSPAXIFBTOHY-ZCFIWIBFSA-N &                   (2R)-2-amino-4-methylpentan-1-ol &      0.48 &         0.02 & 0.19 &  0.02 \\
 98998-25-5 & HJGMRAKQWLKWMH-UHFFFAOYSA-N   &          8-methyl-8-azabicyclo[3.2.1]octan-3-amine &      0.44 &         0.00 & 0.25 &  0.00 \\
   109-83-1 & OPKOKAMJFNKNAS-UHFFFAOYSA-N   &                             2-(methylamino)ethanol &      0.47 &         0.00 & 0.08 &  0.00 \\
   108-13-4 & WRIRWRKPLXCTFD-UHFFFAOYSA-N   &                                     propanediamide &      0.02 &         0.00 & 0.17 &  0.00 \\
   105-59-9 & CRVGTESFCCXCTH-UHFFFAOYSA-N   &             2-[2-hydroxyethyl(methyl)amino]ethanol &      0.24 &         0.02 & 0.00 &  0.00 \\
   141-43-5 & HZAXFHJVJLSVMW-UHFFFAOYSA-N   &                                     2-aminoethanol &      0.55 &         0.04 & 0.09 &  0.01 \\
  6284-40-8 &   MBBZMMPHUWSWHV-BDVNFPICSA-N & (2R,3R,4R,5S)-6-(methylamino)hexane-1,2,3,4,5-p... &      0.71 &         0.03 & 0.20 &  0.01 \\
   110-91-8 & YNAVUWVOSKDBBP-UHFFFAOYSA-N   &                                         morpholine &      0.55 &         0.07 & 0.09 &  0.00 \\
  2752-17-2 & GXVUZYLYWKWJIM-UHFFFAOYSA-N   &                        2-(2-aminoethoxy)ethanamine &      0.37 &         0.12 & 0.13 &  0.00 \\
   497-25-6 &   IZXIZTKNFFYFOF-UHFFFAOYSA-N &                               1,3-oxazolidin-2-one &      0.02 &         0.00 & 0.07 &  0.01 \\
  4847-93-2 &   MECNWXGGNCJFQJ-UHFFFAOYSA-N &                   3-piperidin-1-ylpropane-1,2-diol &      1.32 &         0.23 & 0.00 &  0.00 \\
  2213-43-6 &   LWMPFIOTEAXAGV-UHFFFAOYSA-N &                                  piperidin-1-amine &      0.07 &         0.01 & 0.09 &  0.00 \\
   110-89-4 &   NQRYJNQNLNOLGT-UHFFFAOYSA-N &                                         piperidine &      0.93 &         0.07 & 0.08 &  0.01 \\
  6859-99-0 &   BIWOSRSKDCZIFM-UHFFFAOYSA-N &                                     piperidin-3-ol &      0.28 &         0.11 & 0.29 &  0.08 \\
   110-85-0 &   GLUUGHFHXGJENI-UHFFFAOYSA-N &                                         piperazine &      0.47 &         0.03 & 0.18 &  0.03 \\
  1484-84-0 &   PTHDBHDZSMGHKF-UHFFFAOYSA-N &                            2-piperidin-2-ylethanol &      1.06 &         0.17 & 0.14 &  0.01 \\
  3433-37-2 &   PRAYXGYYVXRDDW-UHFFFAOYSA-N &                             piperidin-2-ylmethanol &      1.30 &         0.11 & 0.10 &  0.00 \\
 22990-77-8 & RHPBLLCTOLJFPH-UHFFFAOYSA-N   &                          piperidin-2-ylmethanamine &      0.52 &         0.00 & 0.42 &  0.12 \\
  5469-70-5 &   LETVJWLLIMJADE-UHFFFAOYSA-N &                                  pyridazin-3-amine &      0.02 &         0.00 & 0.38 &  0.00 \\
  1619-34-7 &   IVLICPVPXWEGCA-UHFFFAOYSA-N &                      1-azabicyclo[2.2.2]octan-3-ol &      1.24 &         0.12 & 0.00 &  0.00 \\
 66211-46-9 & KQIGMPWTAHJUMN-GSVOUGTGSA-N   &                       (2R)-3-aminopropane-1,2-diol &      0.50 &         0.04 & 0.08 &  0.00 \\
  2799-16-8 & HXKKHQJGJAFBHI-GSVOUGTGSA-N   &                            (2R)-1-aminopropan-2-ol &      0.32 &         0.02 & 0.24 &  0.11 \\
 20439-47-8 &   SSJXIUAHEKJCMH-PHDIDXHHSA-N &                    (1R,2R)-cyclohexane-1,2-diamine &      0.38 &         0.02 & 1.20 &  0.07 \\
  2799-17-9 & HXKKHQJGJAFBHI-VKHMYHEASA-N   &                            (2S)-1-aminopropan-2-ol &      0.28 &         0.01 & 0.23 &  0.06 \\
   534-03-2 & KJJPLEZQSCZCKE-UHFFFAOYSA-N   &                            2-aminopropane-1,3-diol &      0.65 &         0.02 & 0.10 &  0.02 \\
 74111-21-0 &   PQMCFTMVQORYJC-WDSKDSINSA-N &                     (1S,2S)-2-aminocyclohexan-1-ol &      0.43 &         0.03 & 0.23 &  0.01 \\
   100-97-0 & VKYKSIONXSXAKP-UHFFFAOYSA-N   &          1,3,5,7-tetrazatricyclo[3.3.1.13,7]decane &      0.01 &         0.00 & 0.08 &  0.00 \\
  5807-14-7 & FVKFHMNJTHKMRX-UHFFFAOYSA-N   & 3,4,6,7,8,9-hexahydro-2H-pyrimido[1,2-a]pyrimidine &      0.30 &         0.01 & 0.35 &  0.05 \\
  4620-70-6 &   IUXYVKZUDNLISR-UHFFFAOYSA-N &                         2-(tert-butylamino)ethanol &      1.09 &         0.14 & 0.08 &  0.02 \\
   102-71-6 & GSEJCLTVZPLZKY-UHFFFAOYSA-N   &                2-[bis(2-hydroxyethyl)amino]ethanol &      0.20 &         0.01 & 0.05 &  0.01 \\
   112-57-2 & FAGUFWYHJQFNRV-UHFFFAOYSA-N   & N'-[2-[2-(2-aminoethylamino)ethylamino]ethyl]et... &      0.39 &         0.00 & 0.00 &  0.00 \\
   110-95-2 &   DMQSHEKGGUOYJS-UHFFFAOYSA-N &           N,N,N',N'-tetramethylpropane-1,3-diamine &      0.66 &         0.17 & 0.00 &  0.00 \\
  4543-96-8 &   SORARJZLMNRBAQ-UHFFFAOYSA-N &               N,N',N'-trimethylpropane-1,3-diamine &      0.58 &         0.02 & 0.14 &  0.01 \\
 15521-18-3 &   PBKGYWLWIJLDGZ-UHFFFAOYSA-N &                       2-(dimethylamino)propan-1-ol &      0.58 &         0.34 & 0.01 &  0.00 \\
   621-56-7 &   LTACQVCHVAUOKN-UHFFFAOYSA-N &                   3-(diethylamino)propane-1,2-diol &      0.96 &         0.03 & 0.11 &  0.01 \\
   140-82-9 &   VKBVRNHODPFVHK-UHFFFAOYSA-N &                  2-[2-(diethylamino)ethoxy]ethanol &      1.07 &         0.13 & 0.12 &  0.01 \\
  4402-32-8 & BHUXAQIVYLDUQV-UHFFFAOYSA-N   &                        1-(diethylamino)propan-2-ol &      1.05 &         0.05 & 0.00 &  0.00 \\
   111-26-2 & BMVXCPBXGZKUPN-UHFFFAOYSA-N   &                                      hexan-1-amine &      0.73 &         0.00 & 0.12 &  0.00 \\
  3554-62-9 & QVUNMWQRCFABKL-UHFFFAOYSA-N   &                        1-propan-2-ylpiperidin-3-ol &      1.07 &         0.28 & 0.00 &  0.00 \\
 40137-22-2 & WOMTYMDHLQTCHY-UHFFFAOYSA-N   &                    3-(methylamino)propane-1,2-diol &      0.65 &         0.02 & 0.13 &  0.01 \\
 27646-80-6 & LHYBRZAQMRWQOJ-UHFFFAOYSA-N   &                2-methyl-2-(methylamino)propan-1-ol &      0.18 &         0.03 & 0.08 &  0.00 \\
  3529-08-6 &   JMUCXULQKPWSTJ-UHFFFAOYSA-N &                     3-piperidin-1-ylpropan-1-amine &      0.35 &         0.17 & 0.13 &  0.03 \\
   627-35-0 &   GVWISOJSERXQBM-UHFFFAOYSA-N &                             N-methylpropan-1-amine &      0.55 &         0.03 & 0.11 &  0.02 \\
  2160-93-2 &   XHJGXOOOMKCJPP-UHFFFAOYSA-N &         2-[tert-butyl(2-hydroxyethyl)amino]ethanol &      0.65 &         0.06 & 0.00 &  0.00 \\
 26734-09-8 &   FNVOFDGAASRDQY-UHFFFAOYSA-N &                    3-amino-2,2-dimethylpropan-1-ol &      0.55 &         0.02 & 0.21 &  0.00 \\
   108-91-8 &   PAFZNILMFXTMIY-UHFFFAOYSA-N &                                    cyclohexanamine &      0.60 &         0.05 & 0.17 &  0.03 \\
  2842-38-8 &   MGUMZJAQENFQKN-UHFFFAOYSA-N &                         2-(cyclohexylamino)ethanol &      0.95 &         0.09 & 0.13 &  0.03 \\
 19059-68-8 &   PYEWZDAEJUUIJX-UHFFFAOYSA-N &          3-(dimethylamino)-2,2-dimethylpropan-1-ol &      0.28 &         0.00 & 0.02 &  0.00 \\
   108-16-7 &   NCXUNZWLEYGQAH-UHFFFAOYSA-N &                       1-(dimethylamino)propan-2-ol &      0.63 &         0.13 & 0.00 &  0.00 \\
   996-35-0 &   VMOWKUTXPNPTEN-UHFFFAOYSA-N &                         N,N-dimethylpropan-2-amine &      0.35 &         0.03 & 0.14 &  0.02 \\
   622-93-5 &   WKCYFSZDBICRKL-UHFFFAOYSA-N &                        3-(diethylamino)propan-1-ol &      0.55 &         0.15 & 0.05 &  0.00 \\
   623-57-4 &   QCMHUGYTOGXZIW-UHFFFAOYSA-N &                  3-(dimethylamino)propane-1,2-diol &      0.28 &         0.08 & 0.02 &  0.00 \\
 13444-24-1 &   ZNPSUOAGONLMLK-UHFFFAOYSA-N &                              1-ethylpiperidin-3-ol &      0.00 &         0.00 & 0.00 &  0.00 \\
   139-87-7 &   AKNUHUCEWALCOI-UHFFFAOYSA-N &              2-[ethyl(2-hydroxyethyl)amino]ethanol &      0.32 &         0.00 & 0.04 &  0.00 \\
   109-56-8 &   RILLZYSZSDGYGV-UHFFFAOYSA-N &                        2-(propan-2-ylamino)ethanol &      0.93 &         0.06 & 0.11 &  0.02 \\
    75-31-0 &   JJWLVOIRVHMVIS-UHFFFAOYSA-N &                                     propan-2-amine &      0.33 &         0.02 & 0.11 &  0.00 \\
 42055-15-2 &   KRGXWTOLFOPIKV-UHFFFAOYSA-N &                         3-(methylamino)propan-1-ol &      0.54 &         0.01 & 0.21 &  0.01 \\
 20845-34-5 &   HXXJMMLIEYAFOZ-UHFFFAOYSA-N &                   (1-methylpiperidin-2-yl)methanol &      0.25 &         0.00 & 0.00 &  0.00 \\
   120-94-5 &   AVFZOVWCLRSYKC-UHFFFAOYSA-N &                                1-methylpyrrolidine &      0.65 &         0.03 & 0.13 &  0.06 \\
  2508-29-4 &   LQGKDMHENBFVRC-UHFFFAOYSA-N &                                 5-aminopentan-1-ol &      0.57 &         0.03 & 0.19 &  0.02 \\
 16369-21-4 &   BCLSJHWBDUYDTR-UHFFFAOYSA-N &                             2-(propylamino)ethanol &      0.64 &         0.02 & 0.14 &  0.01 \\
   107-10-8 &   WGYKZJWCGVVSQN-UHFFFAOYSA-N &                                     propan-1-amine &      0.31 &         0.00 & 0.16 &  0.02 \\
 34381-71-0 &   VCOJPHPOVDIRJK-LURJTMIESA-N &             [(2S)-1-methylpyrrolidin-2-yl]methanol &      0.27 &         0.00 & 0.00 &  0.00 \\
 13952-84-6 &   BHRZNVHARXXAHW-UHFFFAOYSA-N &                                      butan-2-amine &      0.38 &         0.04 & 0.13 &  0.04 \\
\bottomrule
\end{longtable}\label{tab:data}
\end{landscape}


\includegraphics[width=\linewidth,height=\textheight,keepaspectratio]{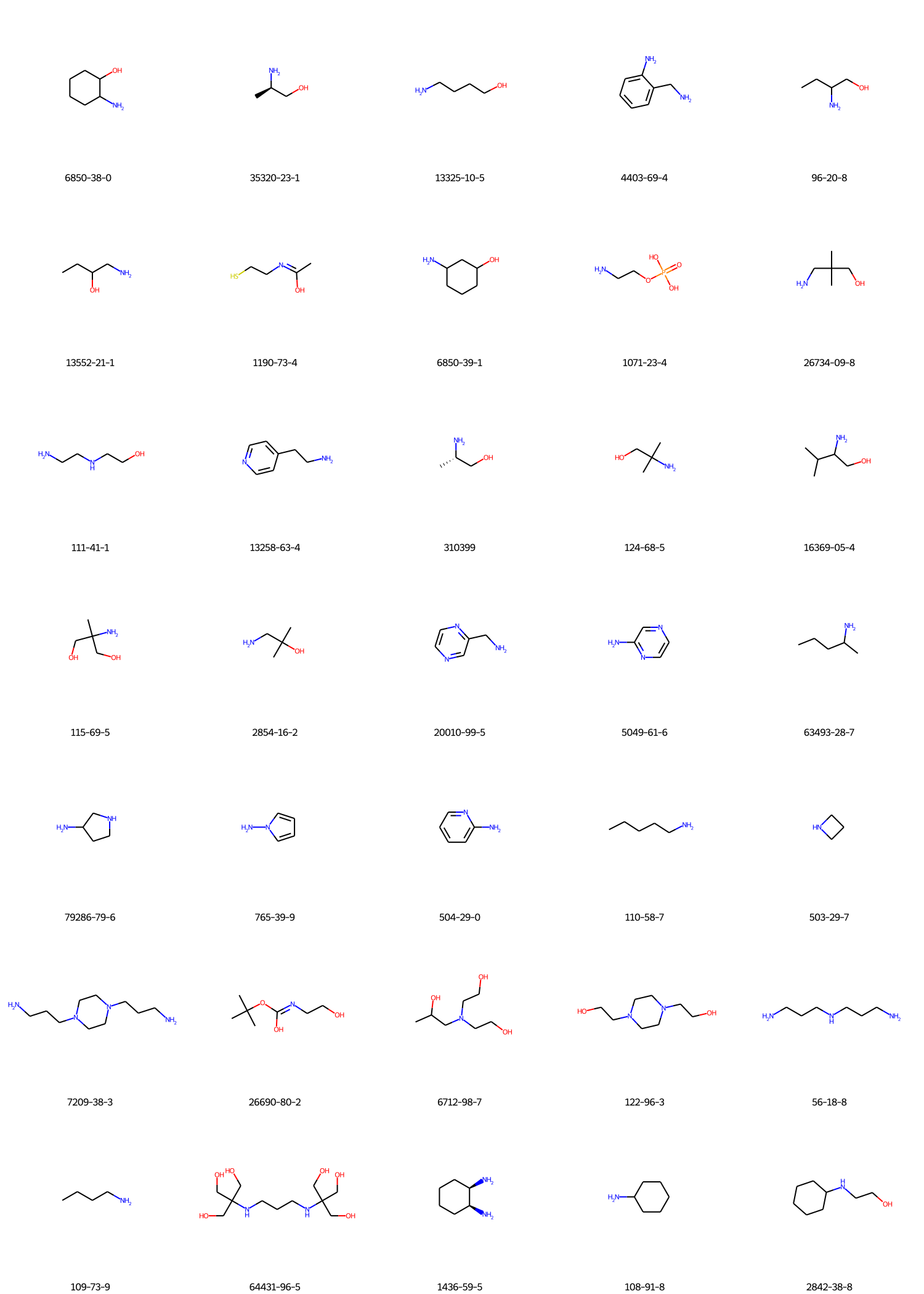}
\newpage
\includegraphics[width=\linewidth,height=\textheight,keepaspectratio]{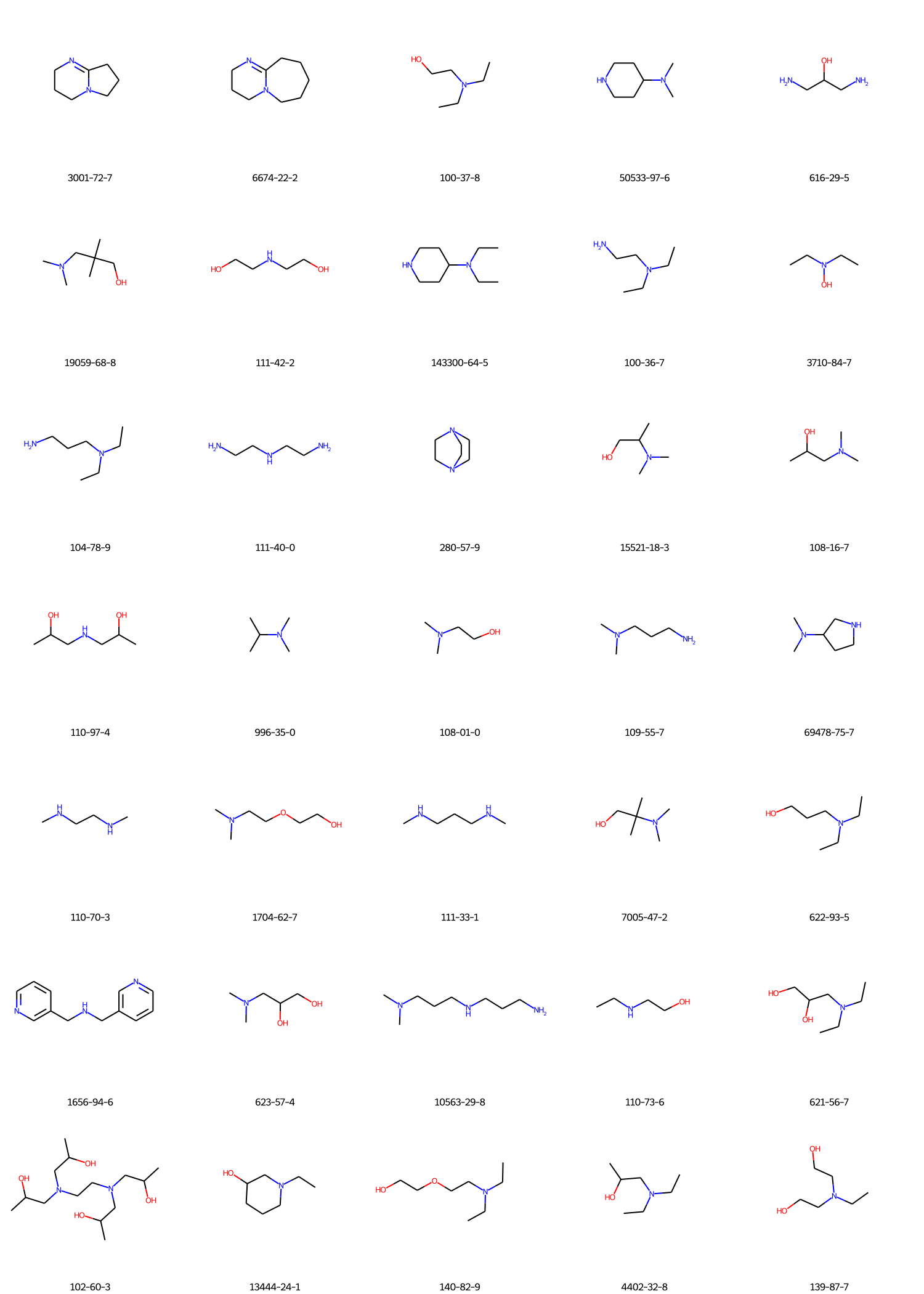}
\newpage
\includegraphics[width=\linewidth,height=\textheight,keepaspectratio]{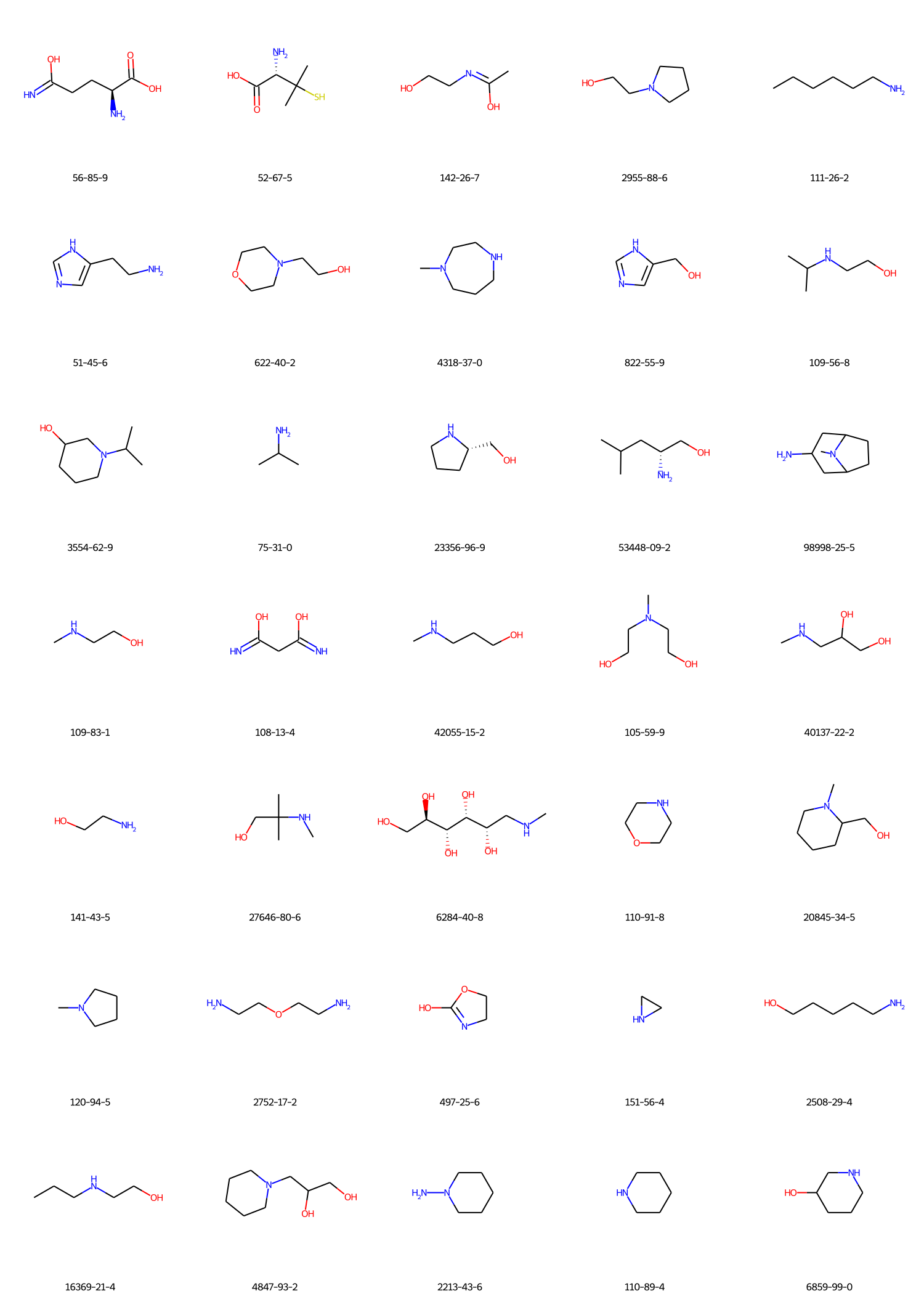}
\newpage
\begin{figure}[H]
\centering
\includegraphics[width=\linewidth,height=\textheight,keepaspectratio]{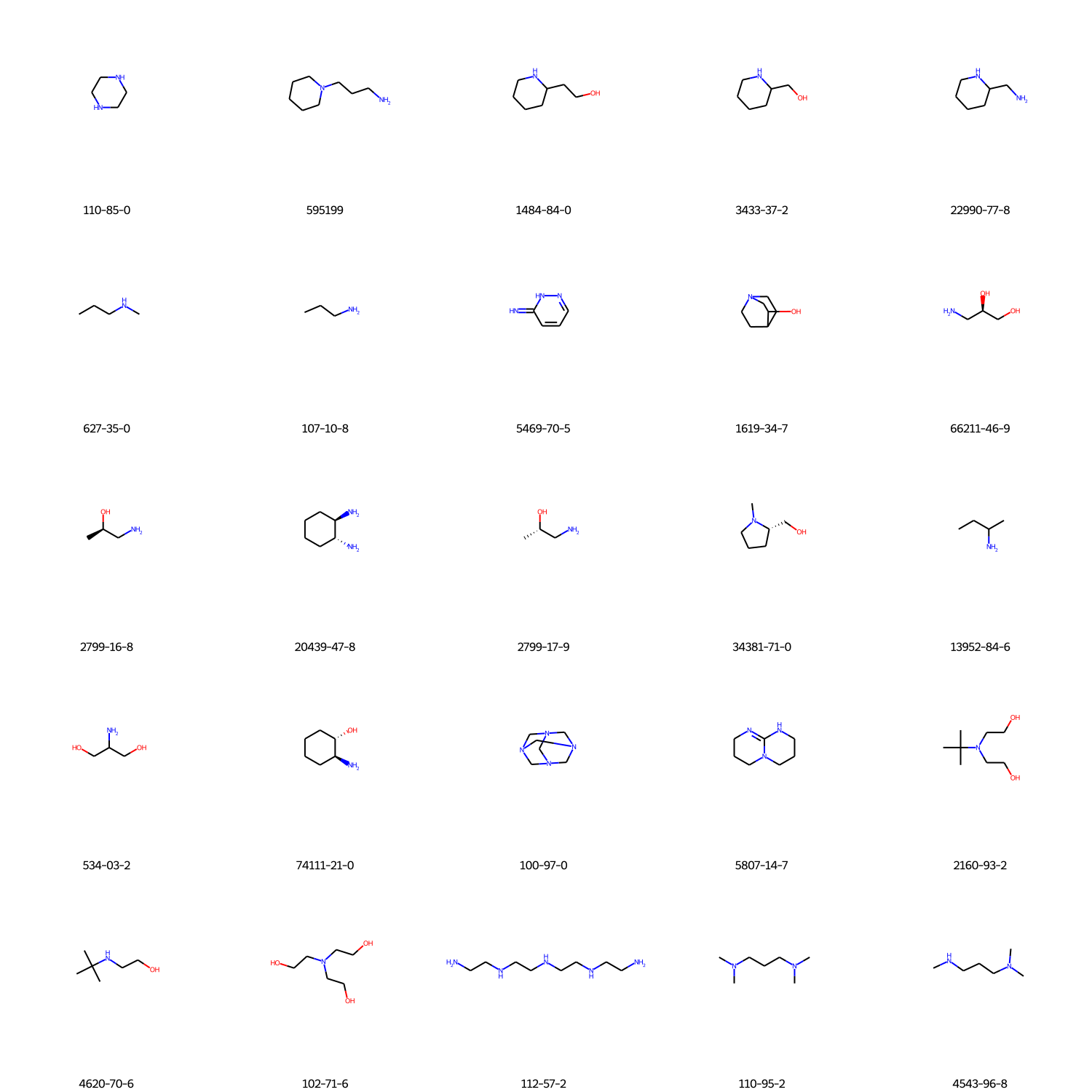}
\caption{Molecular representations of experimentally tested amines }\label{fig:experimental}
\end{figure}

\begin{figure}[H]
\centering
\includegraphics[scale=0.4]{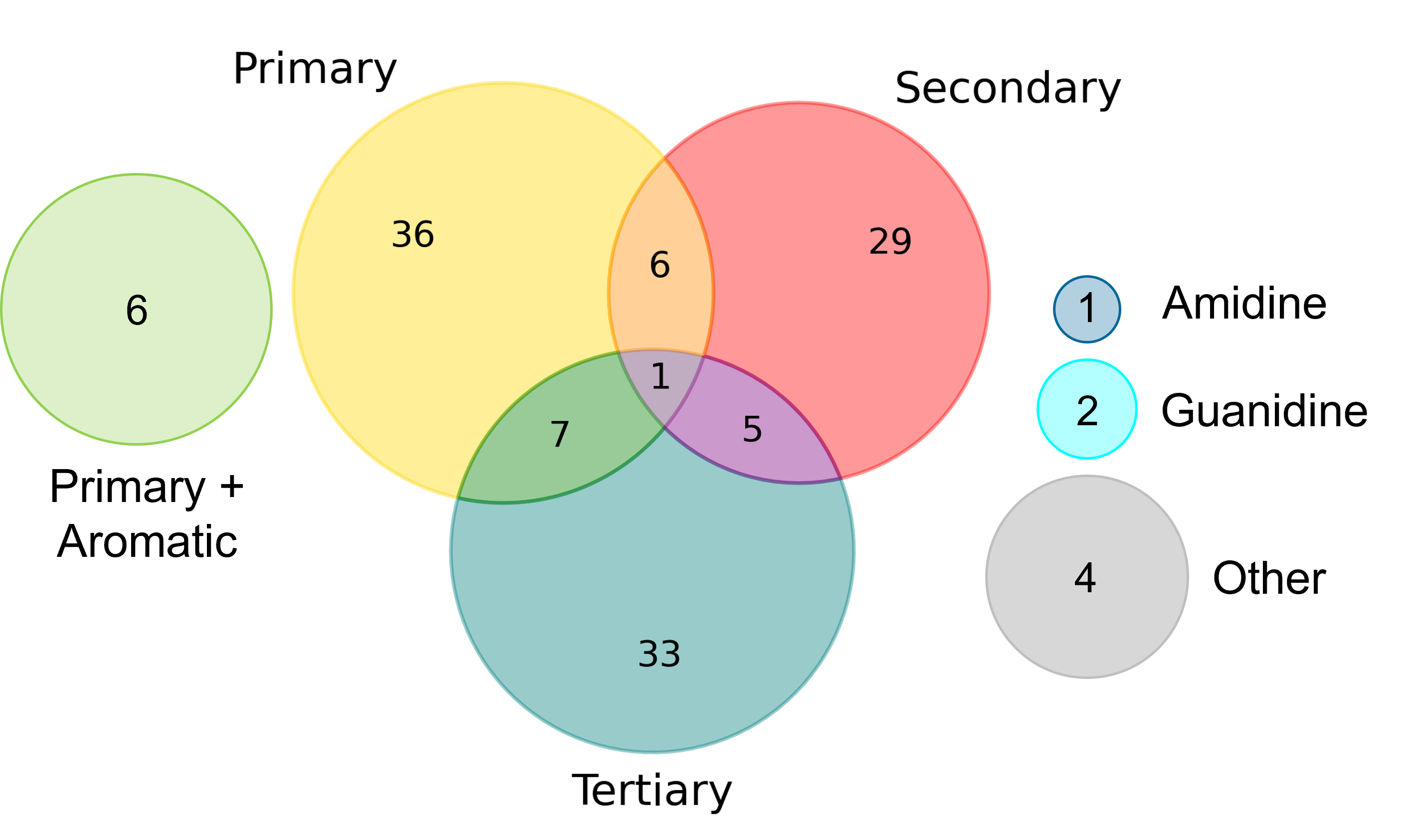}
\caption{Partitioning of amine types within experimental data set }\label{fig:experimental}
\end{figure}

\begin{figure}[H]
\centering
\includegraphics[scale=0.6]{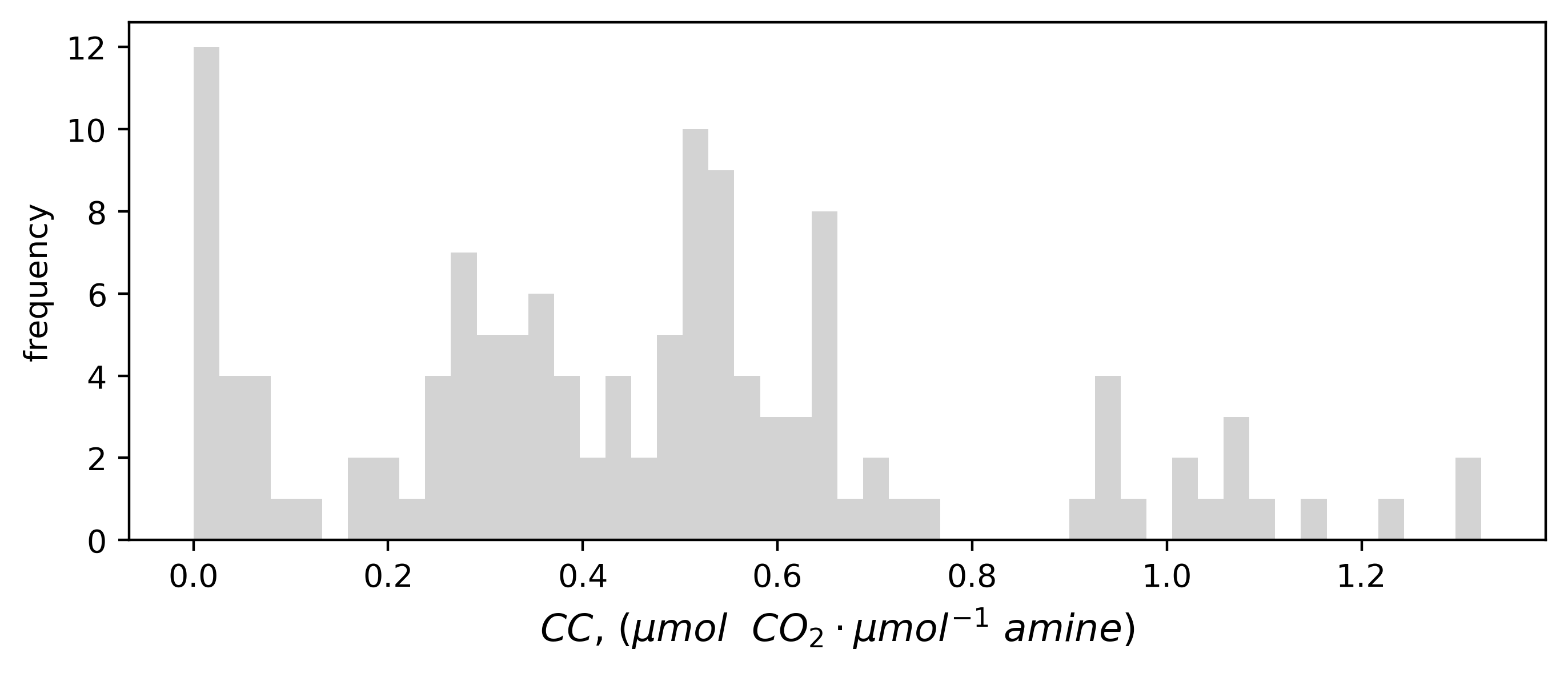}
\caption{Distribution of CO$_{2}$ capture capacities in experimental data set.  50 bins, 0.026 $mol$ $CO_{2} \cdot mol^{-1} amine$ span}\label{fig:experimental}
\end{figure}

\begin{figure}[H]
\centering
\includegraphics[width=0.6\textwidth]{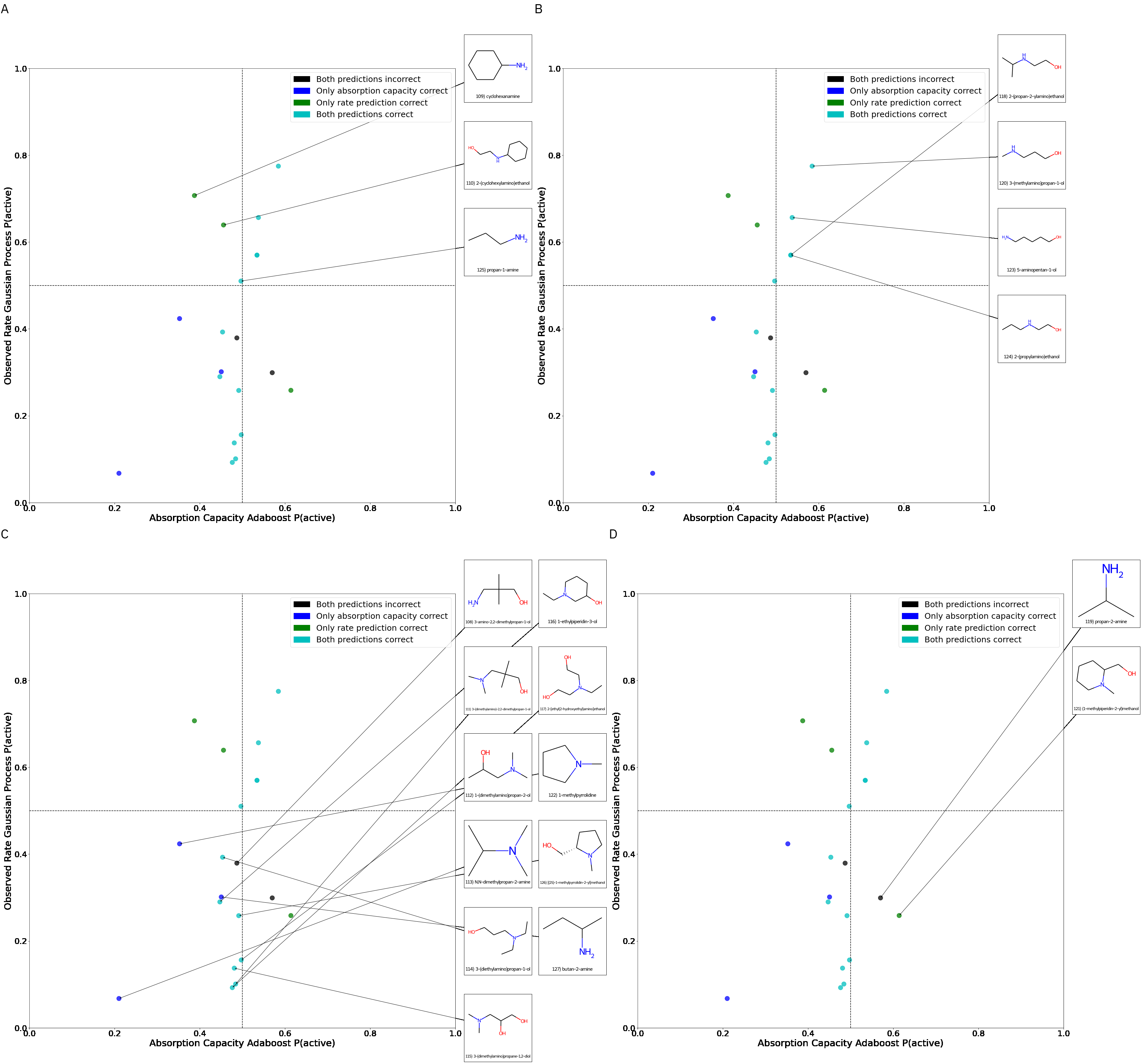}
\caption{Plots of the probability from the classifier model, recapitulating the data from the main text, Figure 4.  The quadrants can be compared directly with those in SI Figure \ref{fig:SI_experimental_plot}}\label{fig:experimental_classes}
\end{figure}

\begin{figure}[H]
\centering
\includegraphics[width=0.6\textwidth]{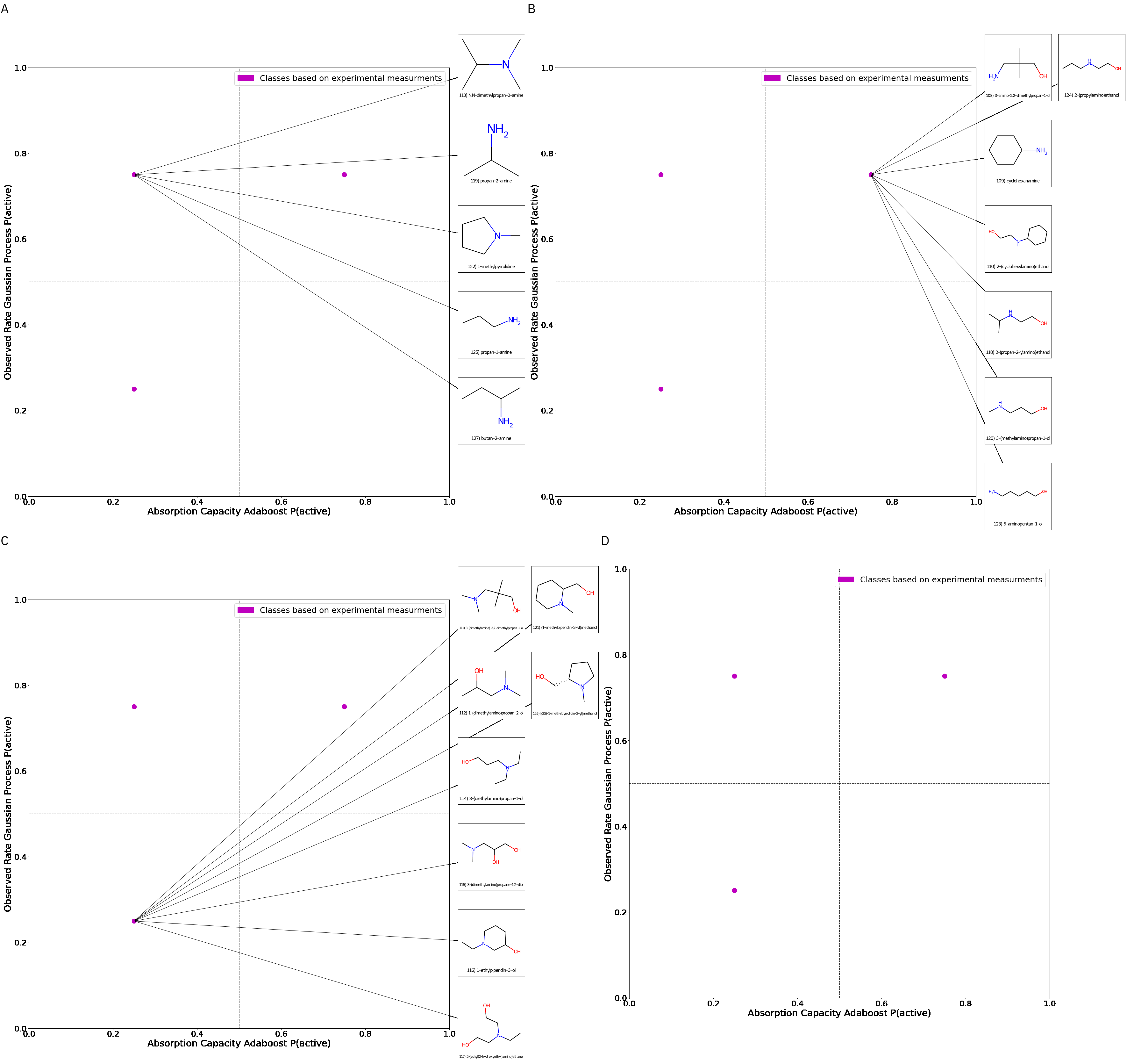}
\caption{Plots showing the actual classes of the amine molecules determined based upon experimental data.}\label{fig:experimental_classes}\label{fig:SI_experimental_plot}
\end{figure}

\newpage

\section{Model Hyper-parameters}\label{sec:hyparams}
\textbf{\emph{\AC{} Models}}:

\noindent \textbf{Decision Tree}:
max depth: 3, max features: None, max leaf nodes: 10, min impurity decrease: 0.0

\noindent \textbf{QDA}: reg param: 0.0

\noindent \textbf{Gaussian Naive Bayes}: priors: 0.67, 0.33

\noindent \textbf{Gaussian Process}: kernel: 1**2 * Matern(length\_scale=0.6, nu=0.5) + WhiteKernel(noise\_level=0.1)

\noindent \textbf{AdaBoost}: learning rate: 0.1, n estimators: 75

\noindent \textbf{DNN}: alpha: 1e-05, batch size: 10, hidden layer sizes: (16, 8), learning rate : constant

\noindent \textbf{Extra Trees}: max depth: 20, max features: sqrt, n estimators: 100

\noindent \textbf{Logistic Regression}: C: 1.0, l1 ratio: 0.5, penalty: elasticnet

\noindent \textbf{k-Nearest Neighbours}: n neighbors: 24, p: 1.5, weights: distance

\noindent \textbf{Support Vector Machine}: C: 0.5, degree: 2, gamma: scale, kernel: rbf

\textbf{\emph{\OIR{} Models}}:

\noindent \textbf{Decision Tree}:
max depth: 3, max features: None, max leaf nodes: 5, min impurity decrease: 0.0

\noindent \textbf{QDA}: reg param: 0.5

\noindent \textbf{Gaussian Naive Bayes}: priors: 0.67, 0.33

\noindent \textbf{Gaussian Process}: kernel: 1**2 * RBF(length scale=1) + WhiteKernel(noise level=0.5)

\noindent \textbf{AdaBoost}: learning rate: 1.25, n estimators: 20

\noindent \textbf{DNN}: alpha: 1e-05, batch size: 5, hidden layer sizes: (72, 36), learning rate: constant

\noindent \textbf{Extra Trees}: max depth: 7, max features: auto, n estimators: 200

\noindent \textbf{Logistic Regression}: C: 0.25, penalty: l2

\noindent \textbf{k-Nearest Neighbours}: n neighbors: 28

\noindent \textbf{Support Vector Machine}: C: 0.2, degree: 2, gamma: scale, kernel: rbf

Grid search was performed over the following space for \ac{}:

\begin{lstlisting}[language=Python]
clf_parameters = [
    {"DecisionTree":{
        "max_depth": [2, 3, 4, 5, 7, 10, 15, 20, 25],
        "max_features": ["auto", "sqrt", "log2", None],
        "max_leaf_nodes": [2, 5, 10, None],
        "min_impurity_decrease": [0.0, 0.1, 0.2, 0.3, 0.4, 0.5]
        }
    },
    {"QDA": {
        "reg_param": [0.0, 0.5, 1.0]
        }
    },
    {"GaussianNB": {
        "priors": [(0.67, 0.33), None],
        }
    },
    {
        "GaussianProcess": {
            "kernel":[
            Matern(length_scale=0.6, 
            length_scale_bounds=(1e-7, 1e7),
            nu=0.5) 
            + WhiteKernel(noise_level=0.1 +
            noise_level_bounds=(0.00000001, 1.0)),
            Matern(length_scale=0.5,
            length_scale_bounds=(1e-7, 1e7),
            nu=1.5) 
            + WhiteKernel(noise_level=0.1,
            noise_level_bounds=(0.00000001, 1.0)),
            Matern(length_scale=0.5,
            length_scale_bounds=(1e-7, 1e7),
            nu=2.5) 
            + WhiteKernel(noise_level=0.1,
            noise_level_bounds=(0.00000001, 1.0)),
            1.0 * RBF(length_scale=1.0),
            1.0 * Matern(length_scale=1.0, nu=1.5),
            1.0 * Matern(length_scale=1.0, nu=2.5),
            1.0 * RBF(1.0) + WhiteKernel(noise_level=0.5)
                     ]
        }
    },
    {
        "AdaBoost": {
                    "n_estimators": [10, 20, 50, 75, 
                    100, 150, 200, 250],
                    "learning_rate": [0.1, 0.5, 0.75,
                    1.0, 1.25, 1.5, 1.7, 2.0]
        }
    },
    {
        "DNN": {
            "alpha": np.logspace(-5, 2, 1), 
            "batch_size": [1, 2, 5, 10],
            "hidden_layer_sizes": [(16),
                                   (32),
                                   (16,8),
                                   (16,8,4)
                                  ],
            "learning_rate": ["constant", "adaptive"]
        }
    },
    {
        "ExtraTreesClassifier": {
            "max_depth": [3, 4, 5, 7, 10, 15, 20], 
            "n_estimators": [50, 100],
            "max_features": [None, "sqrt", "log2"]
        }
    },
    {
        "Logistic_Regression": {
            "penalty": ["l1", "l2", "elasticnet", "none"],
            "C": [0.05, 0.1, 0.25, 0.5, 1.0, 1.25],
            "l1_ratio": [0.1, 0.5, 0.9],
        }
    },
    {
        "Nearest_neighbours": {
            "n_neighbors": [ent for ent in range(2, 25, 1)],
             "p": [1.0, 1.25, 1.5, 1.75, 2.0],
             "weights": ["uniform", "distance"]
        }
    },
    {
        "Support_vector": {
            "kernel": ["linear", "poly", "rbf", "sigmoid"],
            "C": [0.1, 0.2, 0.5, 0.7, 0.9, 
            1.0, 1.2, 1.5, 2.0, 5.0],
            "gamma": ["scale", "auto"],
            "degree": [2, 3, 4, 5]
            
        }
                          
    }
]

\end{lstlisting}

Grid search was performed over the following space for \oir{}:

\begin{lstlisting}[language=Python]
clf_parameters = [
    {"DecisionTree":{
        "max_depth": [2, 3, 4, 5, 7, 10, 15, 20, 25],
        "max_features": ["auto", "sqrt", "log2", None],
        "max_leaf_nodes": [2, 5, 10, None],
        "min_impurity_decrease": [0.0, 0.1, 0.2, 0.3, 0.4, 0.5]
        }
    },
    {"QDA": {
        "reg_param": [0.0, 0.5, 1.0]
        }
    },
    {"GaussianNB": {
        "priors": [(0.67, 0.33), None],
        }
    },
    {
        "GaussianProcess": {
            Matern(length_scale=0.5,
            length_scale_bounds=(1e-7, 1e7),
            nu=1.5) 
            + WhiteKernel(noise_level=0.1,
            noise_level_bounds=(0.00000001, 1.0)),
            Matern(length_scale=0.5,
            length_scale_bounds=(1e-7, 1e7),
            nu=2.5) 
            + WhiteKernel(noise_level=0.1,
            noise_level_bounds=(0.00000001, 1.0)),
            1.0 * RBF(length_scale=1.0),
            1.0 * Matern(length_scale=1.0, nu=1.5),
            1.0 * Matern(length_scale=1.0, nu=2.5),
            1.0 * RBF(length_scale=1) +
            WhiteKernel(noise_level=0.5)
                     ]
        }
    },
    {
        "AdaBoost": {
                    "n_estimators": [10, 20, 50, 75, 
                    100, 150, 200, 250],
                    "learning_rate": [0.1, 0.5, 0.75,
                    1.0, 1.25, 1.5, 1.7, 2.0]
        }
    },
    {
        "DNN": {
            "alpha": np.logspace(-5, 2, 1), 
            "batch_size": [1, 2, 5, 10],
            "hidden_layer_sizes": [(16),
                                   (32),
                                   (16,8),
                                   (16,8,4)
                                  ],
            "learning_rate": ["constant", "adaptive"]
        }
    },
    {
        "ExtraTreesClassifier": {
            "max_depth": [3, 4, 5, 7, 10, 15, 20], 
            "n_estimators": [50, 100],
            "max_features": [None, "sqrt", "log2"]
        }
    },
    {
        "Logistic_Regression": {
            "penalty": ["l1", "l2", "elasticnet", "none"],
            "C": [0.05, 0.1, 0.25, 0.5, 1.0, 1.25],
            "l1_ratio": [0.1, 0.5, 0.9],
        }
    },
    {
        "Nearest_neighbours": {
            "n_neighbors": [ent for ent in range(2, 25, 1)],
             "p": [1.0, 1.25, 1.5, 1.75, 2.0],
             "weights": ["uniform", "distance"]
        }
    },
    {
        "Support_vector": {
            "kernel": ["linear", "poly", "rbf", "sigmoid"],
            "C": [0.1, 0.2, 0.5, 0.7, 0.9, 
            1.0, 1.2, 1.5, 2.0, 5.0],
            "gamma": ["scale", "auto"],
            "degree": [2, 3, 4, 5]
            
        }
                          
    }
]

\end{lstlisting}

\section{Experimental}\label{sec:experiment}
\normalsize
\subsection{Carbon Dioxide Absorption Apparatus}
The CO$_{2}$ absorption apparatus was build in-house.  Gas was fed from a 200L tank (either 100\%v/v N$_{2}$, 100\%v/v CO$_{2}$ (Matheson Gas, Irving, TX, USA), 9.96\%v/v CO$_{2}$ balance N$_{2}$ or 20.0\%v/v CO$_{2}$ balance N$_{2}$ (Airgas, Radnor, PA, USA)) fitted with a regulator, with pressure stepped to 20 psi.  A computer controlled mass flow controller (Sensirion AG, Switzerland) was used to meter gas flow.  Gas was fed at $q = 10$sccm into a 22-gauge steel needle (Sigma Aldrich, St. Louis, MO, USA) which was fitted through a silicon plug (McMaster-Carr, Elmhurst, IL, USA).  The reactor was a NMR tube (Wilmad-Lab Glass, Vineland, NJ, USA) into which 200 $\mu$L sample was injected under nitrogen.  To run, the tube was sealed with the silicon plug and the tube suspended in a pre-heated 40\degree C water bath.  The injection needle end was positioned at the very bottom of the tube.  An outlet was made using a cut off 20-gauge steel needle (insert) mated to a 1/16" OD, 0.03" ID EFTE tube (IDEX Healthcare, Lake Forest, IL, USA) and inserted into the top of the silicon plug.  The outlet needle end sat ~1 cm down from the top of the tube, far away from the reaction mixture and any bubble entertainment.  

Outlet gas was plumbed through ~ 20 cm of the 1/16" OD tubing to a solid-state non-dispersive infra-red detector (NDIR).  The detector was an integrated NDIR sensor + emitter developer kit was purchased from The Kemet Electronics (part number USEQGSK300000, Fort Lauderdale, FL, USA).  This packaged unit could be directly interfaced to a computer for programmed control.  A custom flow cell was 3D printed to give a 1 cm optical wavelength and to interface with the 1/16" OD tubing.  Each run was initiated by synchronizing the gas injection and sensor recording through a single computer interface.  Runs were monitored manually and ended when the IR absorption signal of the outlet gas had saturated to baseline (fraction of CO$_{2}$ in outlet stream, $f_{CO2}$ = injection concentration $f_{o}$.  Runs were typically allowed to go 2x-4x past perceived saturation to ensure the baseline was reached.  The IR absorption was experimentally calibrated each day for no CO$_{2}$ ($f_{CO2}$ = 0) and saturation ($f_{CO2}$ = $f_{o}$) and these calibration values used for analyzing samples day-of.  

\subsection{Signal Analysis}

\begin{figure}[H]
\centering
\includegraphics[scale=0.6]{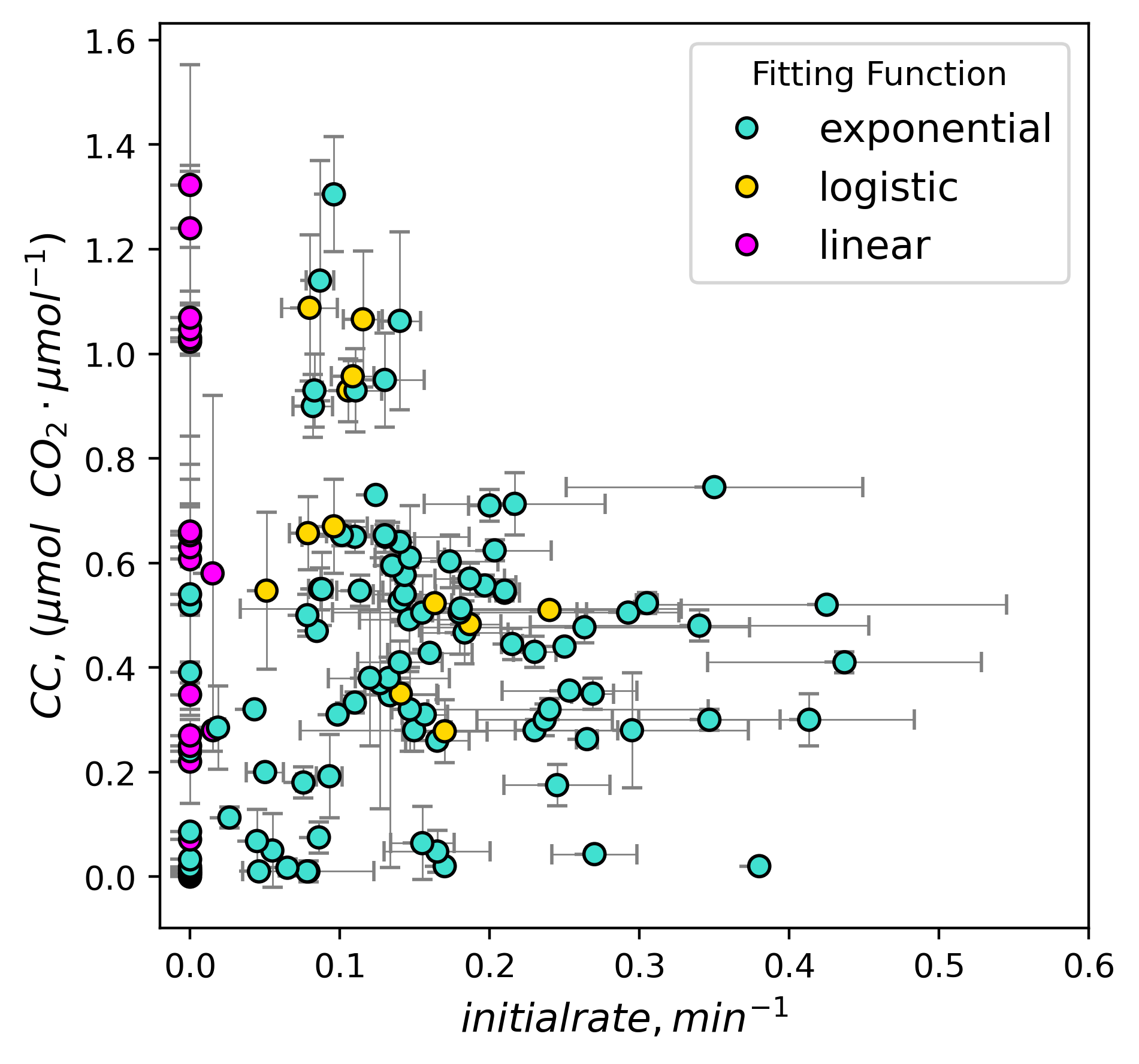}
\caption{Scattering of \ac{} and \oir{} for each experimentally measured amine, coded by the fitting function used to approximate the \coo absorption signal (exponential, logistic or linear).  Distribution of fitting types shows that the slower reacting amines, i.e. \oir{} << 0.1 (predominately tertiary and frustrated secondary amines) required the linear fit, while the faster reacting primary and secondary mainly required exponential fits. }\label{fig:fitting}
\end{figure}

The IR absorption values at 4.3 $\mu$m were converted to transmission, T, by normalizing with the reference signal at 3.9 $\mu$m.  The transmission was normalized by the transmission at zero CO$_{2}$.  In order to properly account for saturation, all normalised transmission curves were approximated with a fitting function, F.  For fast reacting amines, a first-order decaying exponential or logistic function was used, based on the practical issue that some amines exhibited a more rounded signal roll-off that could not be accounted with the exponential form.  The onset of roll-off used to determine the starting point for the fitting function was determined at standard deviation of signal over the mean > 0.005.  When the signal was depleted (i.e. all CO$_{2}$ absorbed, the fitting function in this time span was set to F = 1  For slow reacting amines, the signal was approximated by a linear fit. The choice of exponential or logistic F does not show a strong correlation with the measured metrics (\ref{fig:fitting}).

Absorbance, A, was calculated as \begin{math} A = 1 - F \end{math}. To convert absorbance to CO$_{2}$ a Modified Beer-Lambert equation was used: 
\begin{math}
    C = (ln(1-A/a) / -b])^{c^{-1}}
\end{math} 
Where A is the absorbance, C is the volume fraction concentration of CO$_{2}$ in the outlet stream and $a, b$ and $c$ are fitting parameters.  This was calibrated using $f_{CO2}$ = 0.0, 0.0414 (atmosphere), 10.0 and 20.0 \%v/v CO$_{2}$ at a fixed flow rate of $q =$10 sccm.  Atmospheric CO$_{2}$ was obtained using a house compressed air line.  The CO$_{2}$ absorbed by the amine is C$_{ab}$ = 1 - C.  The volume equivalent of CO$_{2}$ absorbed, V, was calculated using the flow rate.  The cumulative absorbed CO$_{2}$ was obtained by integrating V for each time step.  The cumulative plot was fitted to a first order exponential and the total volume of CO$_{2}$ absorbed extracted.  The total molar equivalent of CO$_{2}$ was estimated using the Ideal Gas Law; from which the capture capacity $\alpha$ could be estimated as the total moles of CO$_{2}$ absorbed over the moles of amine in solution.

\subsection{Stock Solution Preparation and Handling}
All amines were used as received.  Aqueous stock solutions for each amine were prepared in a nitrogen dry box.  Stock solutions were prepared in 20 mL scintillation vials and stored until needed at 4$\degree$C, in the dark, under nitrogen.  For dispensing samples into NMR tubes, a nitrogen shroud was used to blanket the vial while sample was pipetted out.  The sample was then transferred to a clean NMR tube and immediately capped.  Sample sizes were determined from the weight of the tared NMR tube.   

\section{Metrics for classification}

The classic metric for classification is accuracy as calculated in equation \ref{eq:accuracy}. This is an overall metric showing the fraction of correct predictions from all predictions. 

\begin{equation}
\text{accuracy} = \frac{(\text{True Positives} + \text{True Negatives})}{(\text{True Positives} + \text{False Positives} + \text{True Negatives} + \text{False Negatives})}
\end{equation}\label{eq:accuracy}

Sensitivity is calculated as the ratio of the number of true positives divided by the sum of the true positives and the false negatives equation \ref{eq:sensitivity}, and specificity is calculated as the ratio of the number of true negatives divided by the sum of true negatives and false positives equation \ref{eq:specificity}. \\

\begin{equation}
\text{Sensitivity} = \frac{\text{True Positives}}{(\text{True Positives} + \text{False Negatives})}
\end{equation}\label{eq:sensitivity}

\begin{equation}
\text{Specificity} = \frac{\text{True Negatives}}{(\text{True Negatives} + \text{False Positives})}
\end{equation}\label{eq:specificity}

The ROC plot is a useful tool because the curves of different models can be compared directly or indirectly for different thresholds. Also, the area under the curve (AUC) can be used as a summary of the model skill. The shape of the curve contains information about the expected false positive rate, and the false negative rate. Smaller values on the x-axis of the plot indicate lower false positives and higher true negatives, and larger values on the y-axis of the plot indicate higher true positives and lower false negatives. Generally, a model is considered predictive when the curves are shown on the top left of the plot. A perfect model is represented by a line that travels from the bottom left of the plot to the top left and then across the top to the top right and has an AUC of 1. A less predictive classifier is one that cannot discriminate between the classes and would predict a random class or a constant class in all cases. Such a model is represented by a diagonal line from the bottom left of the plot to the top right and has an AUC of 0.5.

Matthews Correlation Coefficient shows the correlation between observed and predicted values and is calculated from equation \ref{eq:mcc}. It returns a value between -1 and 1. A coefficient of 1 represents a perfect prediction and 0 no better than random prediction. Everything above 0.5 is considered a good correlation.
\begin{equation}
MCC = \frac{(TP \times TN – FP \times FN)}{\sqrt{(TP + FP) (TP + FN) (TN + FP)(TN + FN)}}
\end{equation}\label{eq:mcc}

\subsection{Model performance}

\begin{table}[h!]
    \begin{tabular}{lrrrrr}
    
    \toprule
                  Method &  Accuracy &  Sensitivity &  Specificity &  ROC AUC &       MCC \\
    \midrule
            Decision tree &  0.64 &     0.67 &          0.60 & 0.63 &  0.27 \\
                     QDA &  0.45 &     0.50 &          0.40 & 0.45 & -0.10 \\
              Gaussian NB &  0.55 &     0.33 &          0.80 & 0.57 &  0.15 \\
         Gaussian process &  0.73 &     0.83 &          0.60 & 0.72 &  0.45 \\
                Ada Boost &  0.73 &     0.67 &          0.80 & 0.73 &  0.47 \\
                     DNN &  0.64 &     0.67 &          0.60 & 0.63 &  0.27 \\
    Extra Trees &  0.64 &     0.50 &   0.80 & 0.65 &  0.31\\
     Logistic Regression &  0.45 &     0.50 &          0.40 & 0.45 & -0.10 \\
      k-nearest neighbours &  0.64 &     0.83 &          0.40 & 0.62 & 0.26 \\
          Support vector &  0.73 &     0.83 &          0.60 & 0.72 &  0.45 \\
    \bottomrule
    
    \end{tabular}
    \caption{Predictions of \ac{} for the validation set classes using all trained classifiers.}
    \label{tab:capacity_predictions_validation_set}
\end{table}

   \begin{table}[h!]
    \begin{tabular}{lrrrrr}
    \toprule
        Method &  Accuracy &  Sensitivity &  Specificity &  ROC AUC & MCC \\
    \midrule
        Decision Tree &  0.82 &     0.67 &          1.00 & 0.83 & 0.69 \\
        QDA &  0.64 &     0.50 &          0.80 & 0.65 & 0.31 \\
        Gaussian NB &  0.55 &     0.50 &          0.60 & 0.55 & 0.10 \\
        Gaussian process &  0.82 &     0.67 &          1.00 & 0.83 & 0.69 \\
        Ada Boost &  0.73 &     0.67 &          0.80 & 0.73 & 0.47 \\
        DNN &  0.73 &     0.67 & 0.80 & 0.73 & 0.47 \\
        Extra Trees &  0.82 &     0.67 &          1.00 & 0.83 & 0.69 \\
        Logistic regression &  0.73 &     0.67 & 0.80 & 0.73 & 0.47 \\
        k-nearest neighbours &  0.64 &     0.33 & 1.00 & 0.67 & 0.43 \\
        Support vector &  0.73 & 0.67 & 0.80 & 0.73 & 0.47 \\
    \bottomrule
    \end{tabular}
\caption{Predictions of \oir{} for the validation set classes using all trained classifiers.}
    \label{tab:rate_predictions_validation_set}
\end{table}

\begin{table}[h!]
    \centering
    \caption{Table of the predictions from all models in the external test set data. AC is \ac{} and OIR is \oir{}. The \ac{} ensemble is built from models: Decision Tree,  Gaussian Process and Extra Trees Classifier The \oir{} ensemble is built from models: Decision Tree,  Gaussian Process, AdaBoost, DNN, Extra Trees Classifier, Logistic Regression and Support Vector. The ensembles were built using sklearn's Voting Classifier.}
    
    \label{tab:validation_data}
    \begin{tabular}{llrrrrr}
    \toprule
    property &                model &  accuracy &  sensitivity &  specificity &   ROC AUC &    mcc \\
    \midrule
     AC &        DecisionTree &  0.64 &     0.67 &          0.60 & 0.63 &  0.27 \\
     AC &            QDA &  0.45 &     0.50 &          0.40 & 0.45 & -0.10 \\
     AC &     GaussianNB &  0.55 &     0.33 &          0.80 & 0.57 &  0.15 \\
     AC & GaussianProcess &  0.73 &     0.83 &          0.60 & 0.72 &  0.45 \\
    AC &        AdaBoost &  0.73 &     0.67 &          0.80 & 0.73 &  0.47 \\
    AC &             DNN &  0.64 &     0.67 &          0.60 & 0.63 &  0.27 \\
AC & ExtraTreesClassifier &  0.64 &     0.50 &          0.80 & 0.65 &  0.31 \\
 AC & Logistic\_Regression &  0.45 &     0.50 &          0.40 & 0.45 & -0.10\\
  AC & Nearest\_neighbours &  0.64 &     0.83 &          0.40 & 0.62 &  0.26 \\
      AC & Support\_vector &  0.73 &     0.83 &          0.60 & 0.72 &  0.45 \\
      AC & Ensemble & 0.70 & 0.57 & 0.77 & 0.67 & 0.34 \\
         \midrule
    OIR &         DecisionTree &      0.70 &         0.50 &         1.00 &     0.75 & 0.53 \\
       OIR &           QDA &      0.70 &         0.50 &         1.00 &     0.75 & 0.53 \\
      OIR &     GaussianNB &      0.80 &         0.83 &         0.75 &     0.79 & 0.58 \\
     OIR & GaussianProcess &      0.75 &         0.58 &         1.00 &     0.79 & 0.60 \\
     OIR &        AdaBoost &      0.50 &         0.33 &         0.75 &     0.54 & 0.09 \\
     OIR &             DNN &      0.60 &         0.33 &         1.00 &     0.67 & 0.41 \\
OIR & ExtraTreesClassifier &      0.80 &         0.75 &         0.88 &     0.81 & 0.61 \\
 OIR & Logistic\_Regression &      0.75 &         0.58 &         1.00 &     0.79 & 0.60 \\
  OIR & Nearest\_neighbours &      0.60 &         0.33 &         1.00 &     0.67 & 0.41 \\
      OIR & Support\_vector &      0.95 &         0.92 &         1.00 &     0.96 & 0.90 \\
      OIR & Ensemble & 0.65 & 0.42 & 1.00 & 0.71 & 0.47\\
    \bottomrule
    \end{tabular}
\end{table}
\pagebreak
\newpage

\newpage

\section{Other property predictions and performance}
\begin{figure}[h!]
\centering
\includegraphics[width=\textwidth]{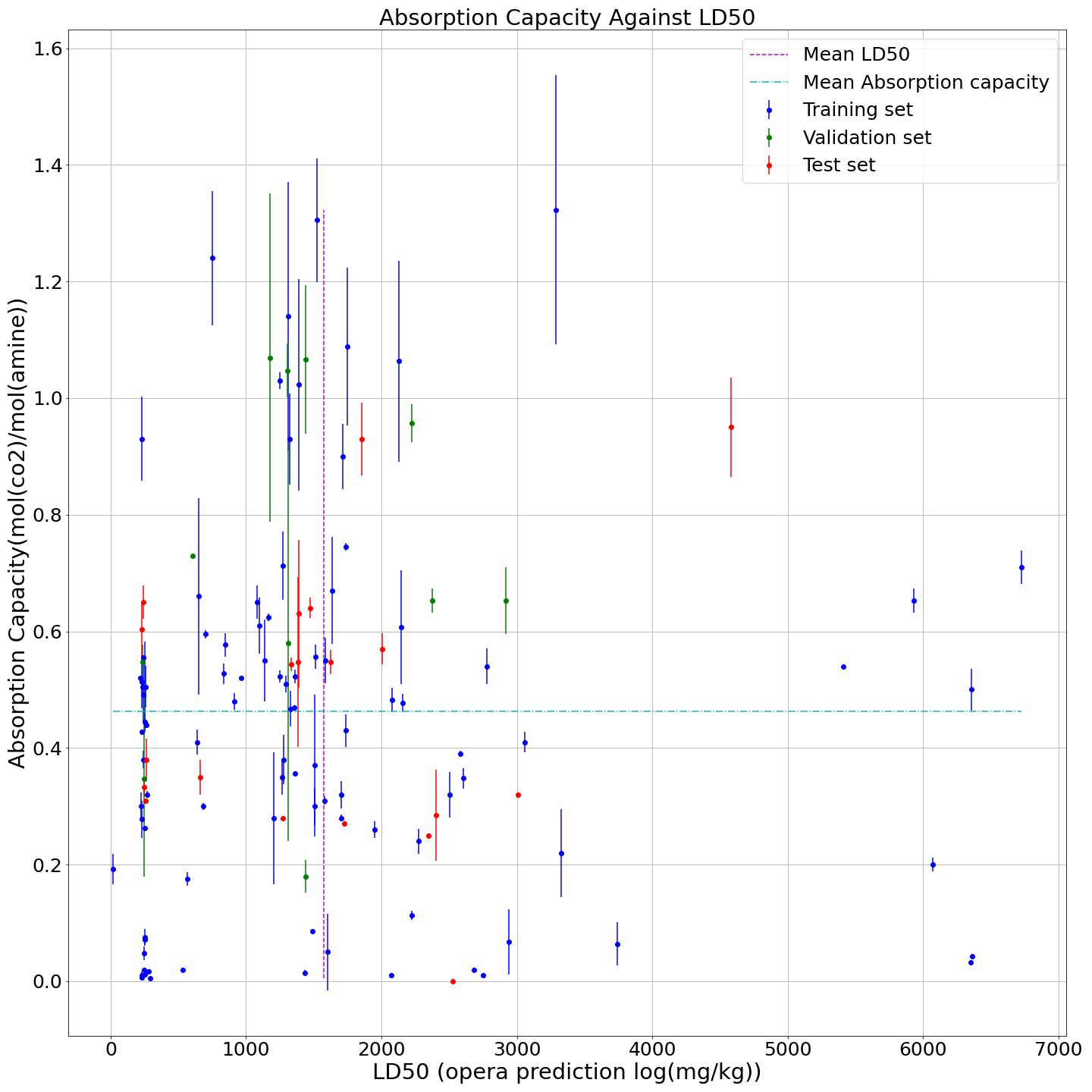}
\caption{\AC{} against the LD50 predicted from the OPERA CATMoS models for toxicity.}\label{fig:cap_vs_ld50}
\end{figure}
\newpage

\begin{figure}[h!]
\centering
\includegraphics[width=\textwidth]{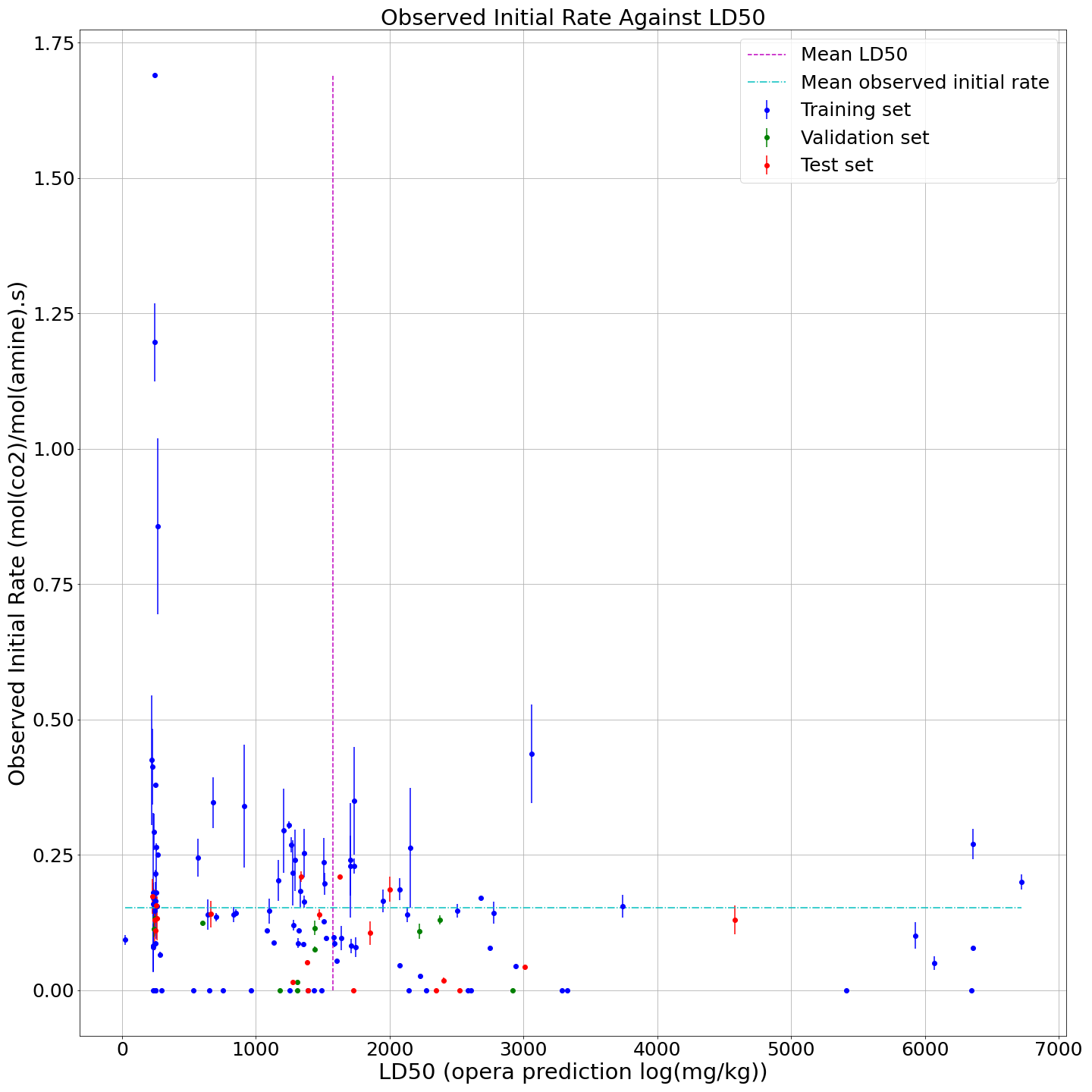}
\caption{\OIR{} against the LD50 predicted from the OPERA CATMoS models for toxicity.}\label{fig:cap_vs_ld50}
\end{figure}
\newpage

\begin{figure}[h!]
\centering
\includegraphics[width=\textwidth]{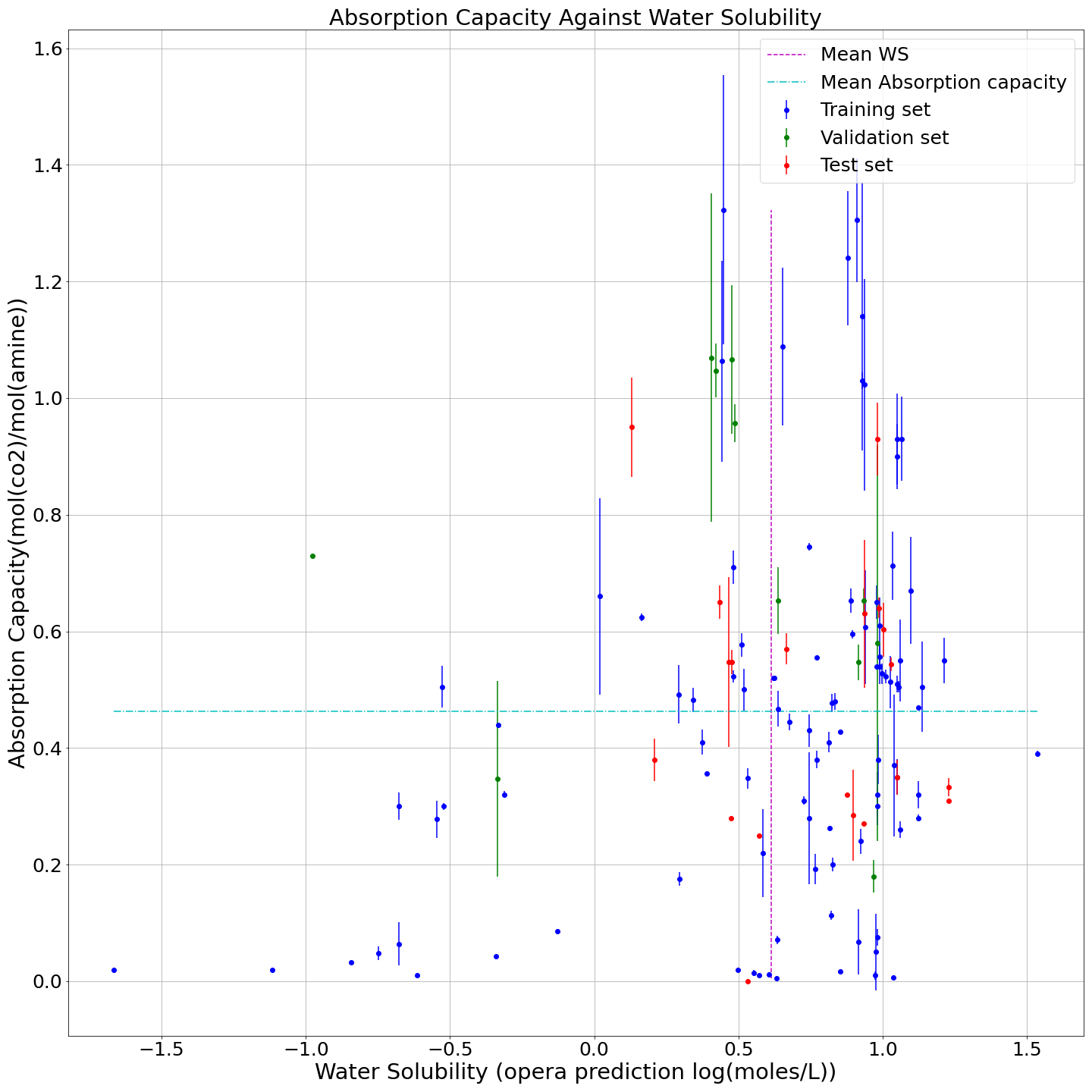}
\caption{\AC{} against the water solubility predicted from the OPERA.}\label{fig:cap_vs_ws}
\end{figure}
\newpage

\begin{figure}[h!]
\centering
\includegraphics[width=\textwidth]{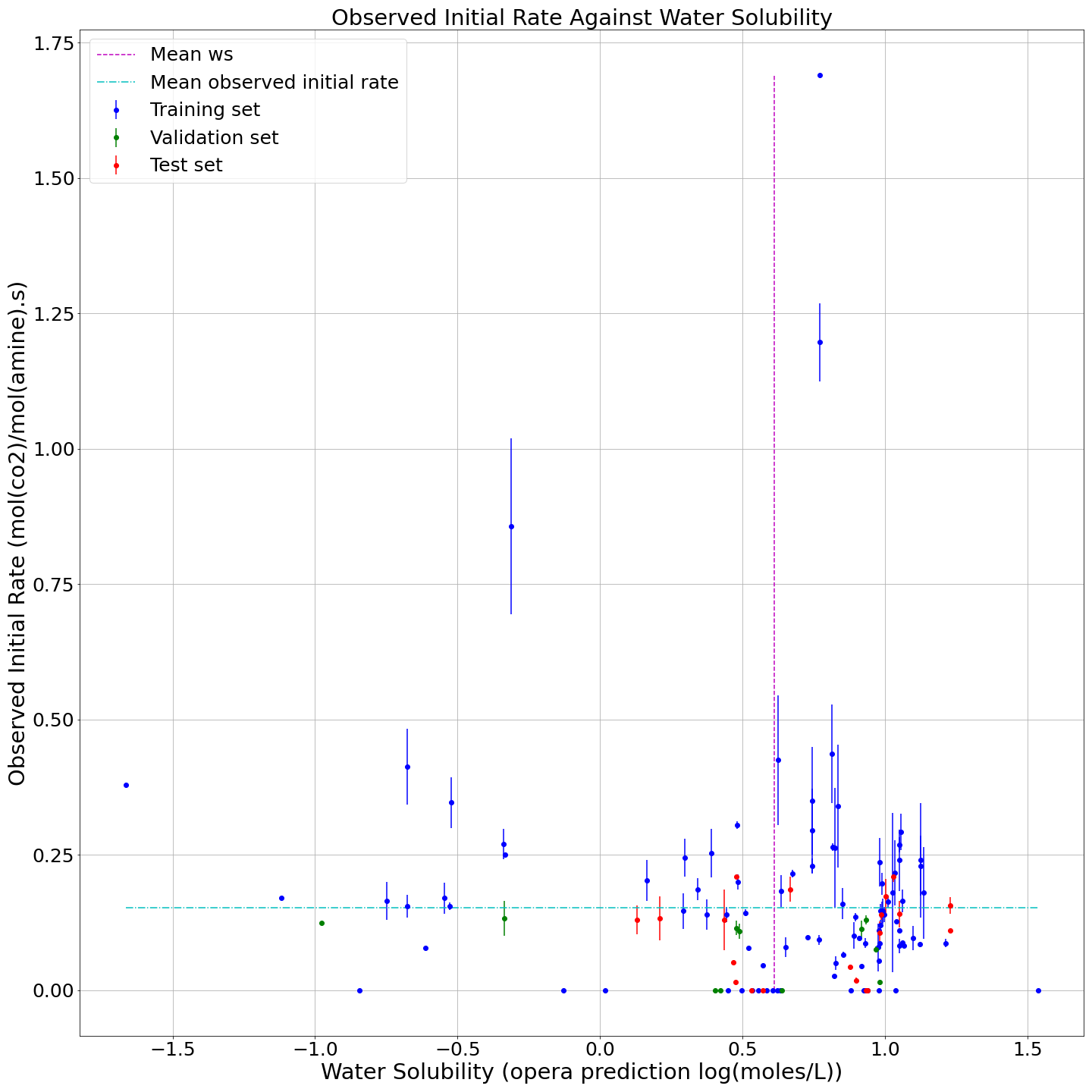}
\caption{\OIR{} against the water solubility predicted from the OPERA.}\label{fig:cap_vs_ws}
\end{figure}
\newpage

\begin{figure}[h!]
\centering
\includegraphics[width=\textwidth]{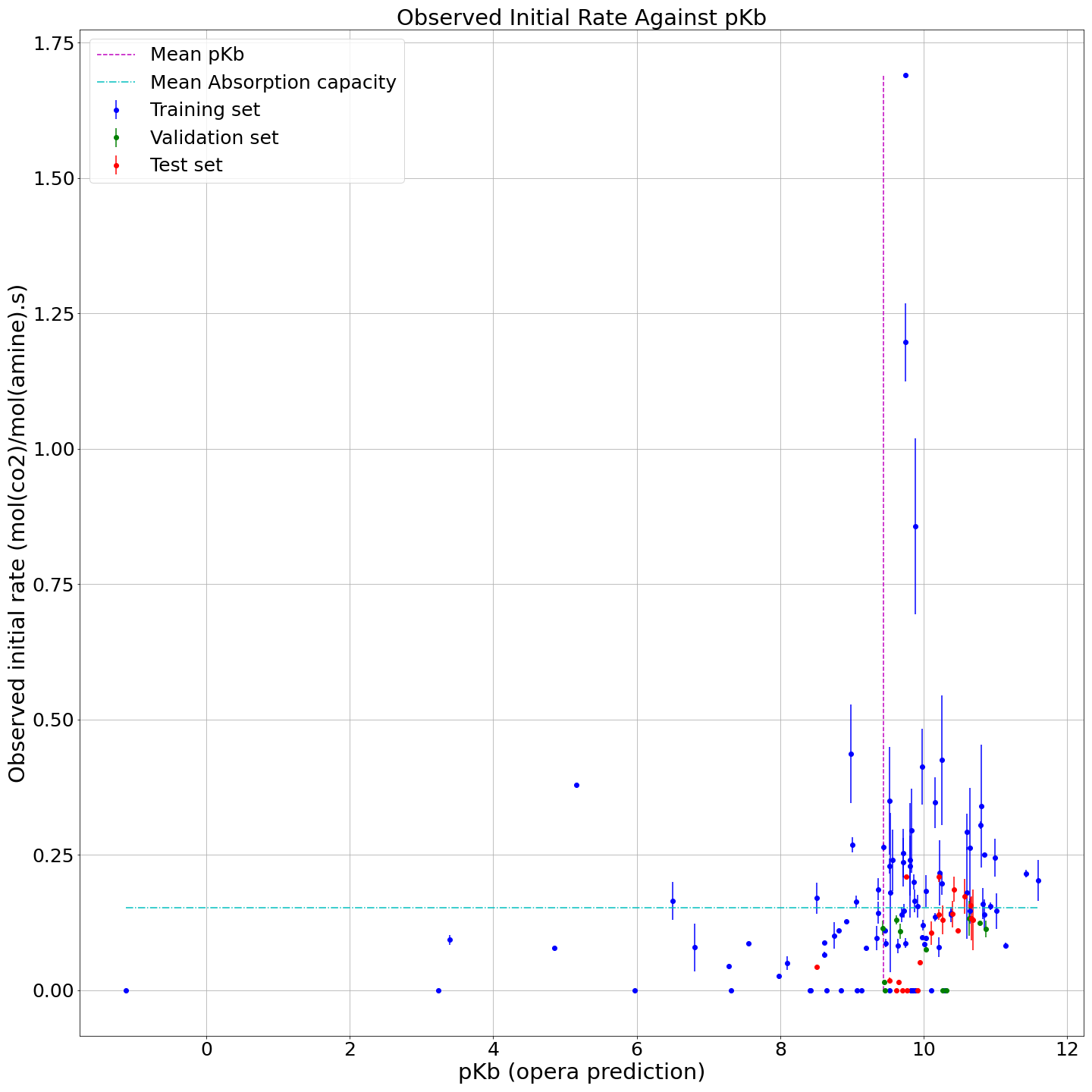}
\caption{\OIR{} against the pKb predicted from the OPERA.}\label{fig:cap_vs_ws}
\end{figure}
\newpage

\begin{figure}[h!]
\centering
\includegraphics[width=\textwidth]{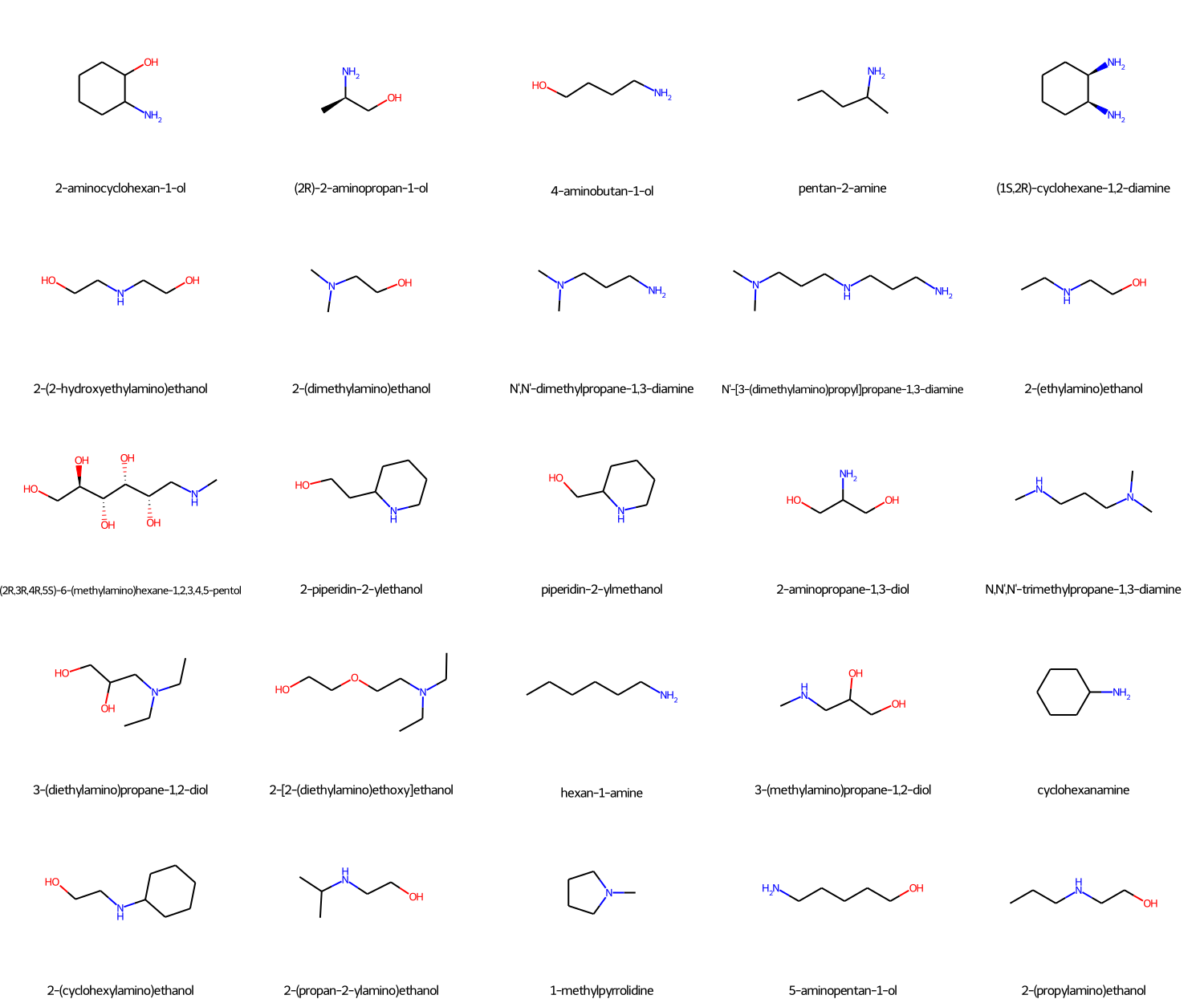}
\caption{Molecules that outperform MEA on both \AC{} and \OIR{}.}\label{fig:better_than_mea}
\end{figure}
\newpage

\begin{figure}[h!]
\centering
\includegraphics[width=\textwidth]{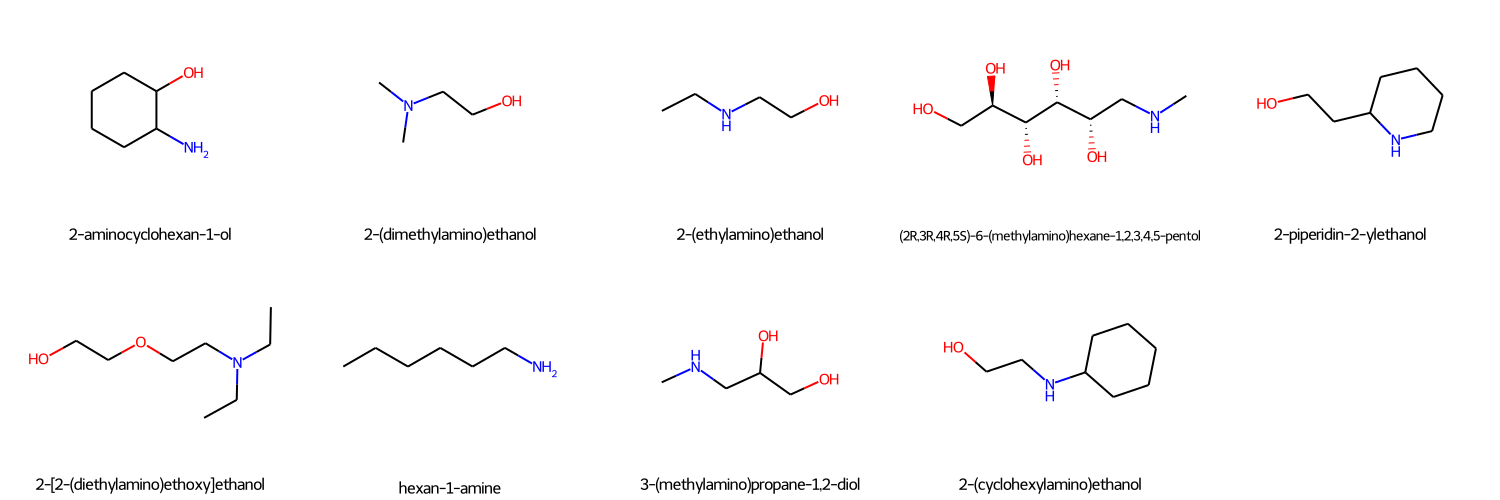}
\caption{Molecules that outperform DEA on both \AC{} and \OIR{}.}\label{fig:better_than_mea}
\end{figure}

\begin{figure}[h!]
    \centering
    \includegraphics[height=\textheight,keepaspectratio]{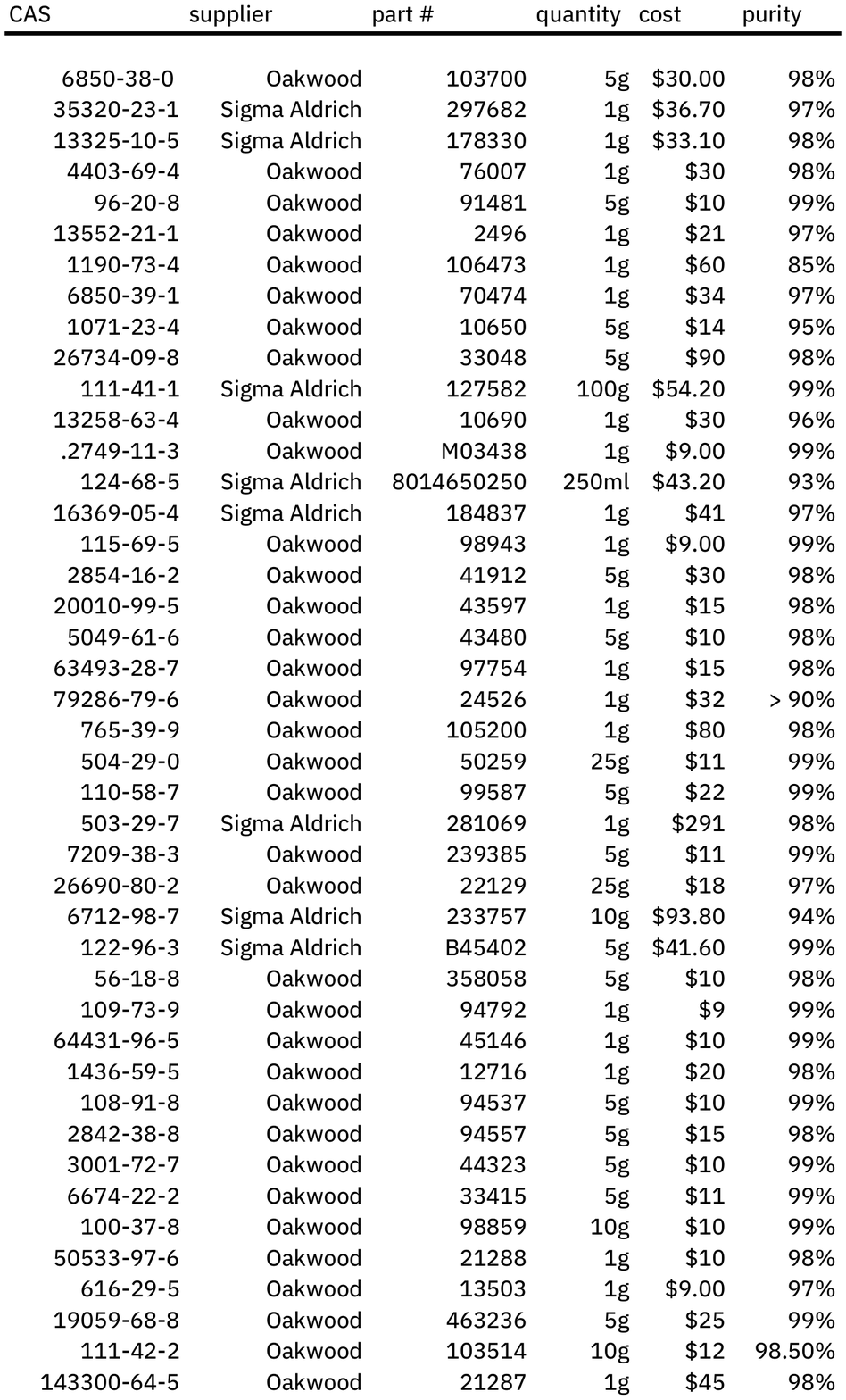}[h!]
\end{figure}

\begin{figure}[h!]
        \includegraphics[height=\textheight,keepaspectratio]{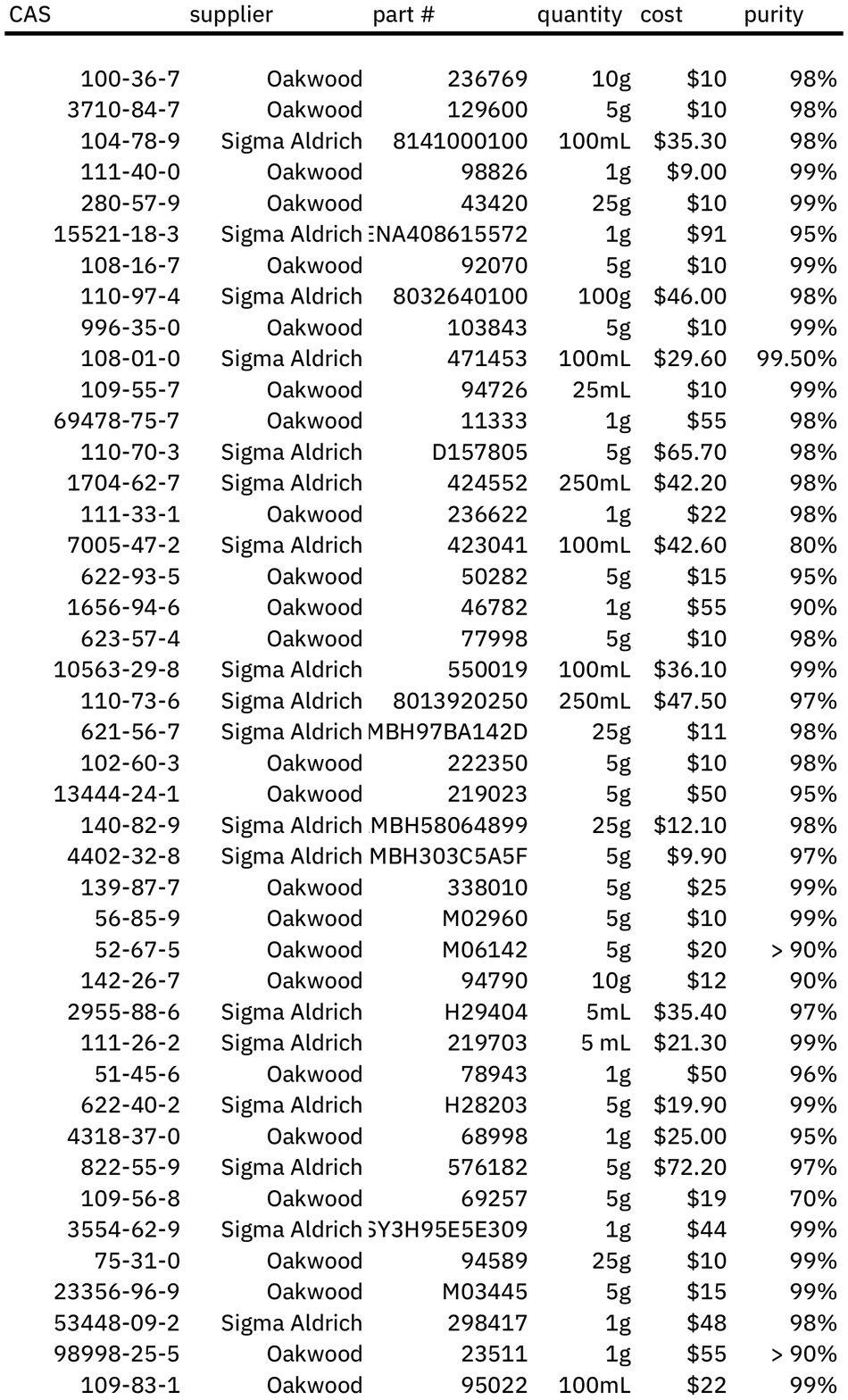}[h!]
\end{figure}

\begin{figure}[h!]
        \includegraphics[height=\textheight,keepaspectratio]{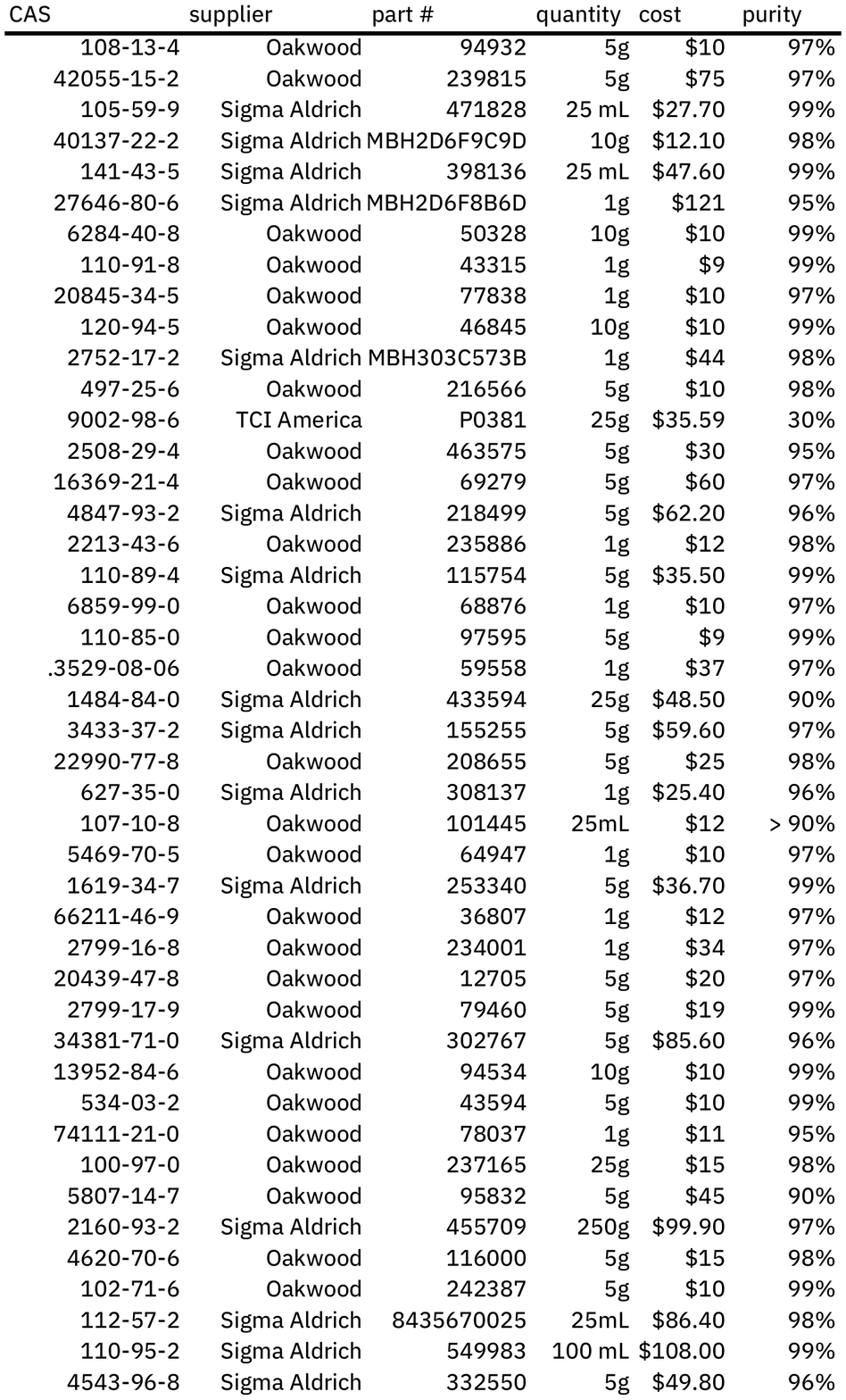}
        \caption{Chemical suppliers, quantities and purity}
\end{figure}